\documentclass[twoside,11pt]{article}

\usepackage[abbrvbib, preprint]{jmlr2e}
\usepackage{xcolor}
\usepackage{booktabs}       
\usepackage{multirow}
\usepackage{graphicx}
\usepackage{caption}
\usepackage{subcaption}
\usepackage{pgfplots}
\usepackage{pgfplotstable}
\usetikzlibrary[patterns]
\usepackage[inline]{enumitem}
\usepackage{amsmath}
\pgfplotsset{compat=1.3}
\usepackage{verbatim}

\newcommand{\wvz}{wav2vec-U}
\newcommand{\wvpp}{wav2vec 2.0}

\newcommand{\wvpplarge}{\textsc{Large}}

\newcommand{\vox}{Libri-Light}
\newcommand{\libri}{Librispeech}

\newcommand{\voxsz}{LL-60k}
\newcommand{\librisz}{LS-960}

\newcommand{\Inp}{\mathcal{X}}
\newcommand{\Feat}{\mathcal{Z}}
\newcommand{\QFeat}{\mathcal{Q}}
\newcommand{\Phonr}{\mathcal{P}^r}

\newcommand{\Context}{\mathcal{C}}
\newcommand{\Seg}{\mathcal{S}}

\newcommand{\E}{\mathop{\mathbb{E}}}

\newcommand{\x}{X}
\newcommand{\ze}{z}
\newcommand{\zq}{q}
\newcommand{\zqt}{\tilde{q}}
\newcommand{\s}{S}

\newcommand{\cc}{c}
\newcommand{\id}{i}
\newcommand{\sss}{s}
\newcommand{\p}{P}
\newcommand{\pr}{P^r}
\newcommand{\pbest}{\hat{\p}}
\newcommand{\pp}{p}
\newcommand{\prand}{\tilde{P}}

\newcommand{\gen}{$\mathcal{G}$}
\newcommand{\genm}{\mathcal{G}}
\newcommand{\dis}{$\mathcal{C}$}
\newcommand{\dism}{\mathcal{C}}

\newcommand{\insertlibrispeech}{
\begin{table*}[h]
\begin{center}
\resizebox{1.0\linewidth}{!}{
\begin{tabular}[b]{lccrrrr}
\toprule
\multirow{2}{*}{Model} & Unlabeled & \multirow{2}{*}{LM} & \multicolumn{2}{c}{dev} & \multicolumn{2}{c}{test} \\
\cline{4-5}\cline{6-7} 
{} & data & {} & clean & other & clean & other \\
\midrule
\midrule
\multicolumn{7}{l}{\textbf{960h - Supervised learning}} \\
DeepSpeech 2~\citep{amodei2016deepspeech} & - & 5-gram & - & - & 5.33 & 13.25 \\
Fully Conv~\citep{zeghidour2018w2l} & - & ConvLM & 3.08 & 9.94 & 3.26 & 10.47 \\
TDNN+Kaldi~\citep{xu2018icassp} & - & 4-gram & 2.71 & 7.37 & 3.12 & 7.63 \\
SpecAugment~\citep{park2019specaugment} & - & - & - & - & 2.8 & 6.8 \\
SpecAugment~\citep{park2019specaugment} & - & RNN & - & - & 2.5 & 5.8 \\
ContextNet~\citep{han2020contextnet} & - & LSTM & 1.9 & 3.9 & 1.9 & 4.1 \\
Conformer~\citep{gulati2020conformer} & - & LSTM & 2.1 & 4.3 & 1.9 & 3.9 \\
\midrule
\multicolumn{7}{l}{\textbf{960h - Self and semi-supervised learning}} \\
Transf. + PL~\citep{synnaeve2020end} & \voxsz{} & CLM+Transf. &  2.00 & 3.65 & 2.09 & 4.11 \\
IPL~\citep{xu2020iterative} & \voxsz{} & 4-gram+Transf. & 1.85 & 3.26 & 2.10 & 4.01 \\
NST~\citep{park2020improved} & \voxsz{} & LSTM & 1.6 & 3.4 & 1.7 & 3.4 \\
\wvpp{}~\citep{baevski2020wav} & \voxsz{} & Transf. & 1.6 & 3.0 & 1.8 & 3.3 \\
\wvpp{} + NST~\citep{zhang2020pushing} & \voxsz{} & LSTM & 1.3 & 2.6 & 1.4 & 2.6 \\
\midrule
\multicolumn{7}{l}{\textbf{Unsupervised learning}} \\
\wvz{} \wvpplarge{} & \voxsz{} & 4-gram & 13.3 & 15.1 & 13.8 & 18.0 \\
\wvz{} \wvpplarge{} + ST & \voxsz{} & 4-gram & 3.4 & 6.0 & 3.8  & 6.5 \\
& \voxsz{} & Transf. & 3.2 & 5.5 & 3.4 & 5.9 \\
\bottomrule
\end{tabular}
}
\caption{
WER on the \libri{} dev/test sets when using 960 hours of unlabeled audio data from \libri{} (\librisz{}) or 53.2k hours from \vox{} (\voxsz{}) using representations from \wvpp{} 
\wvpplarge{}. \libri{} provides clean dev/test sets which are less challenging than the other sets.
We report results for GAN training only (\wvz{}) and with subsequent self-training (\wvz{} + ST).
\label{tab:libri}
}
\end{center}
\vspace{-0.4cm}
\end{table*}
}

\newcommand{\insertTIMIT}{
\begin{table*}[t]
\begin{center}
\resizebox{0.8\linewidth}{!}{
\begin{tabular}[b]{lcrrr}
\toprule
Model & LM & core-dev & core-test & all-test \\
\midrule
\midrule
\multicolumn{5}{l}{\textbf{Supervised learning}} \\
LiGRU~\citep{ravanelli2018lgru} & - & - & 14.9 & - \\
LiGRU~\citep{ravanelli2019pytorchkaldi} & - & - & 14.2 & - \\
\midrule
\multicolumn{5}{l}{\textbf{Self and semi-supervised learning}} \\
vq-wav2vec~\citep{baevski2019vqwav2vec} & - & 9.6 & 11.6 & - \\
\wvpp{}~\citep{baevski2020wav} & - & 7.4 & 8.3 & - \\
\midrule
\midrule
\multicolumn{5}{l}{\textbf{Unsupervised learning - matched setup}} \\
EODM~\citep{yeh2018unsupervised} & 5-gram & - & 36.5 & - \\
GAN$^*$~\citep{chen2019completely} & 9-gram & - & - & 48.6 \\
GAN + HMM$^*$~\citep{chen2019completely} & 9-gram & - & - & 26.1 \\
\midrule[0.3pt]
\wvz{} & 4-gram & 17.0 & 17.8 & 16.6 \\
\wvz{} + ST & 4-gram & 11.3 & 12.0 & 11.3 \\
\midrule
\midrule
\multicolumn{5}{l}{\textbf{Unsupervised learning - unmatched setup}} \\
EODM~\citep{yeh2018unsupervised} & 5-gram & - & 41.6 & - \\
GAN$^*$~\citep{chen2019completely} & 9-gram & - & - & 50.0 \\
GAN + HMM$^*$~\citep{chen2019completely} & 9-gram & - & - & 33.1 \\
\midrule[0.3pt]
\wvz{}$^*$ & 4-gram & 21.3 & 22.3 & 24.4 \\
\wvz{} + ST$^*$ & 4-gram & 13.8 & 15.0 & 18.6 \\
\bottomrule
\end{tabular}
}
\caption{TIMIT Phoneme Error Rate (PER) in comparison to previous work for the matched and unmatched training data setups (\autoref{sec:datasets}). 
PER is measured on the standard Kaldi dev and test sets (core-dev/core-test) as well as a slightly larger version of the test set (all-test) as used by some of the prior work.
$(^*)$ indicates experiments that do not use the standard split excluding SA utterances.
\label{tab:timit}}
\end{center}
\vspace{-0.4cm}
\end{table*}
}

\newcommand{\insertTIMITBoundary}{
\begin{table}[t]
\centering
\resizebox{0.9\linewidth}{!}{
\begin{tabular}{l|rrr}
\toprule
Method & Precision & Recall & F1 \\
\midrule\midrule
DAVEnet + peak detection~\citep{harwath2019towards}
& .893 & .712 & .792 \\
CPC + peak detection~\citep{kreuk2020self} 
& .839 & .836 & .837 \\
\midrule
k-means on \wvpp{} features & .935 & .379 & .539 \\
\wvz{} Viterbi prediction & .598 & .662 & .629 \\
\bottomrule
\end{tabular}
}
\caption{Quantitative evaluation of segment boundaries with respect to human labeled segment boundaries. 
We report precision, recall and f-measure using a 20ms tolerance.
}
\label{tab:boundary}
\end{table}
}

\newcommand{\insertTIMITselftrain}{
\begin{table*}[t]
\begin{center}
\resizebox{0.75\linewidth}{!}{
\begin{tabular}[b]{lcrrr}
\toprule
Model & LM & core-dev & core-test & all-test \\
\midrule
\midrule
\wvz{} & 4-gram & 17.0 & 17.8 & 16.6 \\
~+ HMM & 4-gram & 13.7 & 14.6 & 13.5 \\
~+ HMM + HMM & 4-gram & 13.3 & 14.1 & 13.4\\
~+ HMM resegment + GAN & 4-gram & 13.6 & 14.4 & 13.8 \\
~+ fine-tune & 4-gram & 12.0 & 12.7 & 12.1 \\
\midrule
~+ fine-tune & - & 12.1 & 12.8 & 12.0 \\
~+ fine-tune + fine-tune & - & 12.0 & 12.7 & 12.0  \\
\midrule
~+ HMM + fine-tune & - & 11.3 & 11.9 & 11.3 \\
~+ HMM + fine-tune & 4-gram & 11.3 & 12.0 & 11.3 \\
\bottomrule
\end{tabular}
}
\caption{PER on TIMIT for various self-training strategies. We compare the performance of just the GAN output (\wvz{}) to one or two iterations of subsequent self-training with an HMM. We contrast this to using the HMM for re-segmenting the audio data as done in prior work~\citep{chen2019completely}. We also consider self-training based on fine-tuning the original \wvpp{} model (fine-tune) in or two self-training iterations~\citep{xu2020selftraining} as well as a combination of HMM and fine-tuning-based self-training.
\label{tab:timit_selftrain}}
\end{center}
\vspace{-0.4cm}
\end{table*}
}

\newcommand{\insertMLS}{
\begin{table*}[t]
\begin{center}
\resizebox{1\linewidth}{!}{
\begin{tabular}[b]{lcc|rrrrrr|r}
\toprule
\multirow{2}{*}{Model} & Labeled & \multirow{2}{*}{LM} & \multirow{2}{*}{de} & \multirow{2}{*}{nl} & \multirow{2}{*}{fr} & \multirow{2}{*}{es} & \multirow{2}{*}{it} & \multirow{2}{*}{pt} & \multirow{2}{*}{Avg} \\
& data used & & & & & & & \\
\midrule
\midrule
\multicolumn{3}{l|}{Labeled training hours (full)} & 2k & 1.6k & 1.1k & 918 & 247 & 161  \\
\midrule
\multicolumn{8}{l}{\textbf{Supervised learning}}\\
\citet{pratap2020mls} & full & 5-gram & 6.49 & 12.02 & 5.58 & 6.07 & 10.54 & 19.49 & 10.0 \\
\midrule
\multicolumn{8}{l}{\textbf{Unsupervised learning}}\\
\wvz{} & 0h & 4-gram & 32.5 & 40.2 & 39.8 & 33.3 & 58.1 & 59.8 & 43.9 \\
\wvz{} + ST & 0h & 4-gram & 11.8 & 21.4 & 14.7 & 11.3 & 26.3 & 26.3 & 18.6 \\
\bottomrule
\end{tabular}
}
\caption{WER on the Multilingual Librispeech (MLS) dataset using representations from the \wvpp{} XLSR-53 model. We consider German (de), Dutch (nl), French (fr), Spanish (es), Italian (it), Portuguese (pt).
\label{tab:mls}}
\end{center}
\vspace{-0.4cm}
\end{table*}
}

\newcommand{\insertLowresourceCV}{
\resizebox{1.1\linewidth}{!}{
\begin{tabular}[t]{lrr}
\toprule
Model & tt & ky  \\
\midrule
\midrule
\multicolumn{3}{l}{\textbf{Supervised learning}}\\
\midrule
\citet{fer2017multilingually} & 42.5 & 38.7 \\
m-CPC~\citep{rivire2020unsupervised} & 42.0 & 41.2 \\
XLSR-53~\citep{conneau2020unsupervised} & 5.1 & 6.1 \\
\midrule
\midrule
\multicolumn{3}{l}{\textbf{Unsupervised learning}}\\
\midrule
\wvz{} & 25.7 & 24.1 \\
\wvz{} + HMM & 13.7 & 14.9 \\
\bottomrule
\end{tabular}
}
\caption{PER for low-resource languages, Tatar (tt) and Kyrgyz (ky). 
\label{tab:lowres_cv}}
}

\newcommand{\insertLowresourceALFFA}{
\resizebox{0.73\linewidth}{!}{
\begin{tabular}[t]{lr}
\toprule
Model & sw  \\
\midrule
\midrule
\multicolumn{2}{l}{\textbf{Supervised learning}}\\
\midrule
\citet{besacier2015speechtf} & 27.36 \\
\midrule
\midrule
\multicolumn{2}{l}{\textbf{Unsupervised learning}}\\
\midrule
\wvz{} & 52.6 \\
\wvz{} + ST & 32.2 \\
\bottomrule
\end{tabular}
}
\caption{WER for Swahili from the ALFFA corpus.
We compare to the supervised baseline of the ALFFA project. 
\label{tab:lowres_alffa}}
}

\newcommand{\insertAblation}{
\begin{table*}[h!]
\begin{center}
\resizebox{0.8\linewidth}{!}{
\begin{tabular}[b]{lcc}
\toprule
Ablation & mean PER $\pm$ std & \%-converged (PER $<$ 40)\\
\midrule
Baseline & 21.4 $\pm$ 1.2 & 100\% \\
\midrule
9.6h audio, 3k text & 21.2 $\pm$ 1.1 & 100\% \\
96h audio, 3k text & 21.1 $\pm$ 1.3 & 95\% \\
\midrule
w/o clustering, pca, mean pool & - & 0\% \\
w/o clustering & - & 0\% \\
w/o 2nd stage mean pool & - & 0\% \\
w/o PCA & - & 0\% \\
\midrule
64 clusters & 23.1 $\pm$ 0.7 & 100\% \\
256 clusters & 22.3 $\pm$ 1.1 & 100\%\\
\midrule
256 PCA & 21.6 $\pm$ 1.1 & 100\% \\
768 PCA & 28.0 $\pm$ 1.5 & 90\% \\
\midrule
use full phone set & 23.51 $\pm$ 1.3 & 100\% \\
\bottomrule
\end{tabular}
}
\caption{Ablation of various data settings, pre-processing steps, cluster sizes, PCA sizes and using the full phoneme set.
\label{tab:ablation}}
\end{center}
\vspace{-0.4cm}
\end{table*}
}

\newcommand{\insertSilence}{
\begin{tabular}{lr}
\toprule
& PER \\
\midrule
\midrule
Baseline & 21.4 $\pm$ 1.2 \\
w/o begin/end SIL tokens & 25.8 $\pm$ 0.7 \\
w/o audio silence removal & 29.3 $\pm$ 2.0 \\ 
\bottomrule
\end{tabular}
}

\pgfplotstableread[row sep=\\,col sep=&]{
layer & en_LS \\
1 & 48.6 \\
2 & 88.1 \\
3 & 87.9 \\
4 & 87.0 \\
5 & 88.4 \\
6 & 24.7 \\
7 & 21.0 \\
8 & 86.6 \\
9 & 86.8 \\
10 & 16.8 \\
11 & 16.4 \\
12 & 15.1 \\
13 & 11.9 \\
14 & 8.9 \\
15 & 7.5 \\
16 & 7.7 \\
17 & 8.1 \\
18 & 8.2 \\
19 & 8.8 \\
20 & 9.6 \\
21 & 11.4 \\
22 & 78.6 \\
23 & 100.0 \\
24 & 96.2 \\
}\layerdatals

\pgfplotstableread[row sep=\\,col sep=&]{
layer & mean & stdv \\
1 & 52.5 & 15.5 \\
2 & 54.9 & 20.5 \\
3 & 54.7 & 26.3 \\
4 & 39.3 & 19.6 \\
5 & 61.5 & 30.7 \\
6 & 44.9 & 31.2 \\
7 & 43.6 & 31.9 \\
8 & 43.1 & 32.4 \\
9 & 43.2 & 33.3 \\
10 & 26.0 & 22.3 \\
11 & 34.3 & 29.3 \\
12 & 35.0 & 30.2 \\
13 & 25.4 & 22.3 \\
14 & 24.5 & 23.0 \\
15 & 15.8 & 5.2 \\
16 & 15.3 & 5.0 \\
17 & 15.4 & 4.9 \\
18 & 16.1 & 5.1 \\
19 & 17.3 & 4.9 \\
20 & 19.5 & 5.0 \\
21 & 21.2 & 4.8 \\
22 & 100.0 & 0.0 \\
23 & 100.0 & 0.0 \\
24 & 100.0 & 0.0 \\
}\layerdatamls

\pgfplotstableread[row sep=\\,col sep=&]{
perc & nhours & mean & stdv \\
1\% & 9.6h & 21.78 & 1.3 \\
10\% & 96h & 20.68 & 1.0 \\
25\% & 240h & 21.18 & 1.6 \\
50\% & 480h & 21.23 & 1.4 \\
100\% & 960h & 21.40 & 1.2 \\
}\audiodata

\pgfplotstableread[row sep=\\,col sep=&]{
perc & nsent & mean & stdv \\
0.001\% & 300 & 23.01 & 2.5 \\
0.01\% & 3k & 21.53 & 1.3 \\
0.1\% & 31k & 21.40 & 1.1 \\
1\% & 0.3m & 21.48 & 0.9 \\
10\% & 3m & 21.56 & 1.2 \\
25\% & 8m & 21.44 & 1.2 \\
50\% & 15m & 21.88 & 1.4 \\
100\% & 31m & 21.40 & 1.2 \\
}\textdata

\pgfplotstableread[row sep=\\,col sep=&]{
perc & mean & stdv \\
0\% & 24.87 & 1.2 \\
15\% & 21.97 & 1.0 \\ 
25\% & 21.40 & 1.2 \\
50\% & 21.79 & 0.9 \\
75\% & 22.72 & 1.0 \\
}\silencetokdata

\pgfplotstableread[row sep=\\,col sep=&]{
setting & lbld & unlblbd \\
core-dev, wav2vec-U & 17.2 & 17.8 \\
core-test, wav2vec-U & 16.3 & 16.6 \\
core-dev, wav2vec-U + ST & 10.9 & 12.0 \\
core-test, wav2vec-U + ST & 10.1 & 11.3 \\
}\metricdata

\firstpageno{1}

\begin{document}

\title{Unsupervised Speech Recognition}

\author{\centering 
Alexei Baevski$^{\bigtriangleup}$, 
Wei-Ning Hsu$^{\bigtriangleup}$, 
Alexis Conneau$^{\Box}$\thanks{Work done while at Facebook AI.}~,
Michael Auli$^{\bigtriangleup}$ \\
{\normalfont $^{\bigtriangleup}$~Facebook AI~~~$^{\Box}$~Google AI\\}
}

\editor{Kevin Murphy and Bernhard Sch{\"o}lkopf}

\maketitle

\begin{abstract}
Despite rapid progress in the recent past, current speech recognition systems still require labeled training data which limits this technology to a small fraction of the languages spoken around the globe.
This paper describes wav2vec-U, short for wav2vec Unsupervised, a method to train speech recognition models without any labeled data.
We leverage self-supervised speech representations to segment unlabeled audio and learn a mapping from these representations to phonemes via adversarial training.
The right representations are key to the success of our method.
Compared to the best previous unsupervised work, wav2vec-U reduces the phoneme error rate on the TIMIT benchmark from 26.1 to 11.3.
On the larger English Librispeech benchmark, wav2vec-U achieves a word error rate of 5.9 on test-other, rivaling some of the best published systems trained on 960 hours of labeled data from only two years ago.
We also experiment on nine other languages, including low-resource languages such as Kyrgyz, Swahili and Tatar.
The code is available at \url{https://github.com/pytorch/fairseq/tree/master/examples/wav2vec/unsupervised}
\end{abstract}


\section{Introduction}

Speech recognition performance on the much studied English \libri{} benchmark~\citep{panayotov2015librispeech} has seen rapid improvement over the last few years due to advances in model architectures~\citep{linhao2018transformer,synnaeve2020end,gulati2020conformer}, semi-supervised learning~\citep{xu2020iterative,park2020improved} and self-supervised learning~\citep{oord2018cpc,chung2018speech2vec,chung2019unsupervised,baevski2020wav}.
However, all of these techniques require transcribed speech data which is not available for the vast majority of the nearly 7,000 languages of the world~\citep{lewis2016ethnologue}. 
As a result, speech recognition technology is only available for about 125 different languages~\citep{google2021asr}.
On the other hand, humans learn a lot about speech simply by listening to others around them and without explicit supervision~\citep{werker1984cross,HIRSHPASEK1987269,polka1994developmental,JUSCZYK1999159,JOHNSON2001548}.

Unsupervised learning has been very successful in machine translation resulting in systems that obtain remarkable accuracy given no labeled training data at all~\citep{conneau2018unsupmt,lample2018unsupmt,artetxe2018unsupervised}.
Inspired by this, there has been some work on unsupervised speech recognition based on learning to align unlabeled text and audio~\citep{yeh2018unsupervised} or adversarial learning~\citep{liu2018completely,chen2019completely}.
These approaches showed promising initial results but their error rates are still high, with evaluation being limited to the small-scale and clean TIMIT benchmark.

\begin{figure}[t]
\centering
\includegraphics[width=1\textwidth]{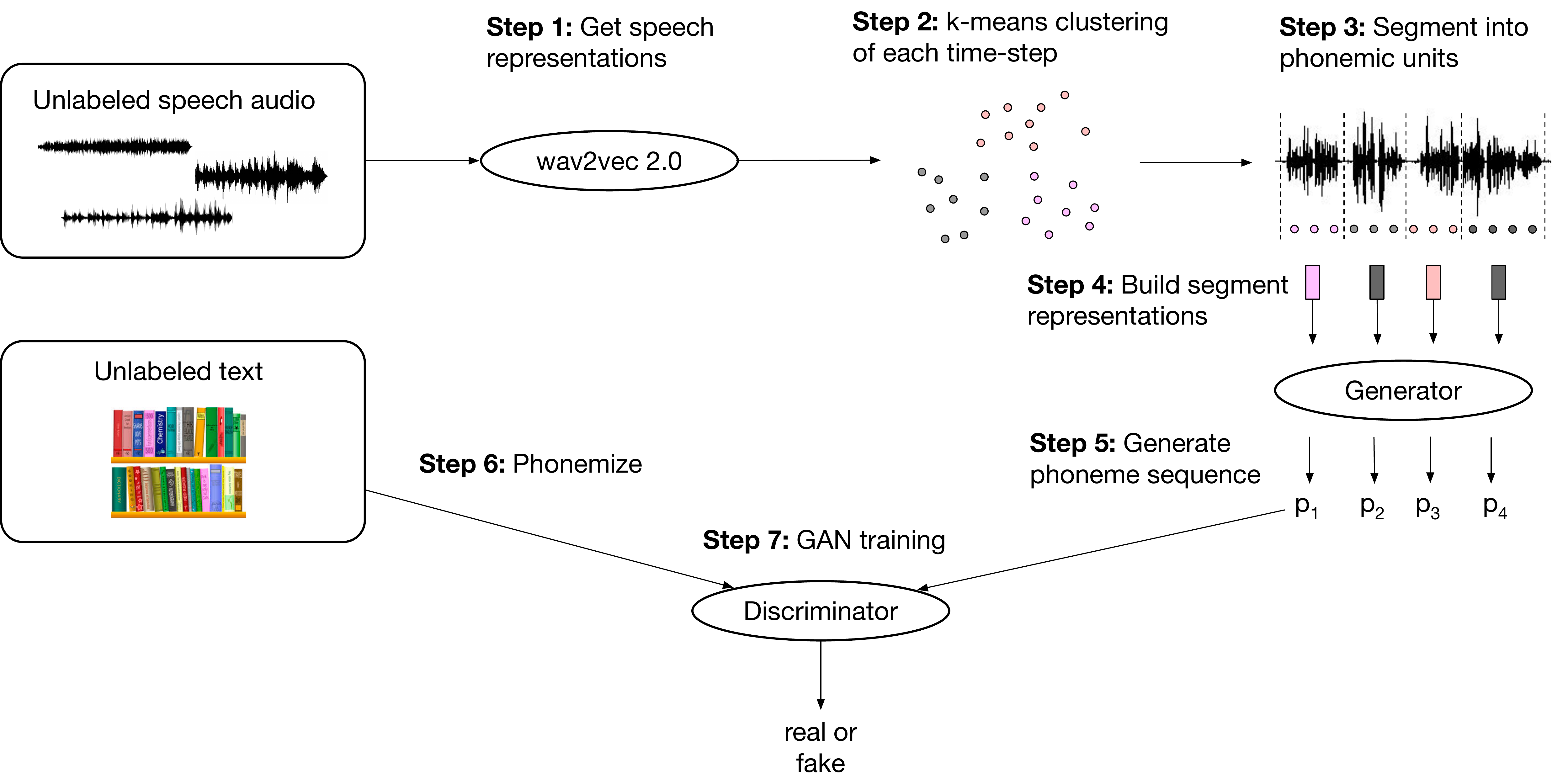} 
\caption{
Illustration of wav2vec Unsupervised: we learn self-supervised representations with \wvpp{} on unlabeled speech audio (Step 1), then identify clusters in the representations with k-means (Step 2) to segment the audio data (Step 3). 
Next, we build segment representations by mean pooling the \wvpp{} representations, performing PCA and a second mean pooling step between adjacent segments (Step 4). 
This is input to the generator which outputs a phoneme sequence (Step 5) that is fed to the discriminator, similar to phonemized unlabeled text (Step 6) to perform adversarial training (Step 7).
}
\label{fig:main}
\end{figure}

In this work, we introduce a framework for unsupervised learning of speech recognition models. Wav2vec-U, or wav2vec Unsupervised, leverages self-supervised representations from \wvpp{}~\citep{baevski2020wav} to embed the speech audio and to segment the audio into units with a simple k-means clustering method (see~\autoref{fig:main} for an illustration of our approach).
We find that the quality of the audio representations is key to the success of unsupervised speech recognition.
Similar to~\citet{liu2018completely}~and~\citet{chen2019completely}, we learn a mapping between segments and phonemes using adversarial training but different to their work, we also enable the algorithm to label segments as silences.
We also introduce an unsupervised cross-validation metric to enable model development without labeled development data.
Our unsupervised speech recognition model, the generator, is very lightweight: it consists of a single temporal convolution comprising only about 90k parameters to which we input frozen \wvpp{} representations.

Experimental results demonstrate the viability of the framework for a variety of settings and languages. 
\wvz{} improves the phone error rate (PER) on the small-scale TIMIT benchmark from 26.1 to 11.3 compared to the next best known unsupervised approach.
To get a better sense of the performance compared to the best supervised methods, we measure performance on the larger \libri{} benchmark where our method achieves word error rate (WER) 5.9 on test-other. 
We also evaluate on six other European languages of the multilingual Librispsech benchmark~\citep{pratap2020mls} and on three non-European low-resource languages.

\section{Background}

\subsection{Supervised ASR}
\label{sec:background_supervised}

Automatic speech recognition (ASR) is the task of transcribing a speech waveform into a text transcript. 
In a supervised setup, a dataset $\mathcal{D}_l = \{(X_i, Y_i)\}_{i=1}^N$ of $N$ speech $X$ and text $Y$ pairs are provided.
Specifically, speech is usually represented as a sequence of feature vectors $X = [x_1, \cdots x_T] \in (\mathbb{R}^m)^*$, where each feature frame $x_t$ is an $m$-dimensional continuous vector such as Mel-frequency cepstral coefficients (MFCC).
Text is usually represented as a sequence of discrete units $Y = [y_1, \cdots, y_L] \in B^*$, where $B$ denotes the vocabulary of text sequences (e.g., letters or words). 
The goal is to build a model, typically a probabilistic one such as $p_\theta(Y \mid X)$, to predict the transcription given speech audio. 
Such models are often classified into two categories: hybrid~\citep{juang1986maximum,young1996large,povey2005discriminative,hinton2012deep,abdel2012applying,bourlard2012connectionist} or end-to-end~\citep{graves2006ctc,graves2012sequence,chorowski2015attention,rao2017exploring}. 

\subsection{Hybrid System for Supervised ASR}
\label{sec:hmm}

Hybrid systems parameterize the \emph{joint distribution} of speech and text $p_\theta(X, Y)$ with three components: an acoustic model (AM), a pronunciation model (PM), and a language model (LM)~\citep{young1996large}. 
A language model describes the distribution over text sequences: $p_{LM}(Y) = \prod_{i=1}^L p(y_i \mid y_1, \cdots, y_{i-1})$. 
A pronunciation model is a lexicon mapping each word to its pronunciation represented as a sequence of phonemes, often created by linguists. For simplicity, we consider a deterministic lexicon here where each word only has one pronunciation, and therefore each word sequence $Y$ can be transcribed into one phoneme sequence $P = [p_1, \cdots, p_M] = f_{PM}(Y) \in O^*$, where $O$ is the phoneme inventory.

The last component, the acoustic model $p_{AM}(X \mid P)$, parameterizes the distribution over speech $X$ given a phoneme sequence $P$. Since each phoneme $p \in O$ can generate a speech feature sequence of variable length, this component is often formulated as a linear hidden Markov model (HMM), where each phoneme has its corresponding HMM model and each state can transition to itself or the next state. 
The probability of a phoneme sequence is described by concatenating the corresponding HMM models (linking the forward transition from the last state of a phoneme to the first state of the next phoneme in that sequence).
Let an alignment $A = [a_1, \cdots, a_T]$ denote a sequence of states from the concatenated HMM each speech frame $x_t$ corresponds to, the joint probability of speech and alignment can be written as
\begin{align*}
     p_{AM}(X, A \mid P) 
        &= p_{emit}(X \mid A)\; p_{tran} (A \mid P) \\
        &= \prod_t p_{emit}(x_t \mid a_t)\; p_{tran} (a_t \mid a_{t-1};\; P),
\end{align*}
where $p_{tran}$ denotes the transition probability derived from the concatenated HMM model, and $p_{emit}$ denotes the emission probability, which is often parameterized by a Gaussian mixture model (GMM).

To compute the acoustic model probability, one can marginalize over all alignments: $p_{AM}(X \mid P) = \sum_A p_{emit}(X \mid A)\; p_{tran} (A \mid P)$ efficiently with the forward-backward algorithm. 
Oftentimes, the probability is approximated with $p_{AM}(X \mid P) \approx \max_A p_{emit}(X \mid A)\; p_{tran} (A \mid P)$, which can also be computed efficiently with the Viterbi algorithm. The resulting alignment, called the \emph{forced alignment}, can be used to segment the speech by phone identities~\citep{chen2019completely} or to create frame-level targets for training neural network acoustic model that parameterizes $p_{emit}(a_t \mid x_t)$~\citep{bourlard2012connectionist}.

\subsubsection{Training a Hybrid System}
Combining AM, PM and LM, a hybrid system computes the joint probability of speech and text as
\begin{align*}
    p_\theta(X, Y) 
        &= p_{AM}(X \mid f_{PM}(Y)) \; p_{LM}(Y)\\
        &= \sum_A p_{emit}(X \mid A) \; p_{tran} (A \mid f_{PM}(Y)) \; p_{LM}(Y) \\
        &\approx \max_A p_{emit}(X \mid A) \; p_{tran} (A \mid f_{PM}(Y)) \; p_{LM}(Y).
\end{align*}
The acoustic and the language model are often learned separately with a maximal likelihood objective~\citep{bahl1983maximum, juang1986maximum}:
\begin{align*}
    p_{LM}^* = \arg\max_{p_{LM}} \sum_{i=1}^N \log p_{LM}(Y_i) \quad\quad
    p_{AM}^* = \arg\max_{p_{AM}} \sum_{i=1}^N \log p_{AM}(X_i \mid f_{PM}(Y_i)),
\end{align*}
but alternative discriminative objectives for AMs taking LMs into account have also been explored~\citep{bahl1986maximum,povey2005discriminative}.
Note that since training of language models requires only text data, one can leverage not only text from the paired data in $\mathcal{D}_l$, but also additional unpaired text data for training.

\subsubsection{Decoding a Hybrid System with Weighted Finite State Transducers}
\label{sec:background_wfst}
Once the hybrid system is trained, decoding amounts to finding the most probable text sequence for a given speech sequence $\arg\max_Y p_\theta(Y \mid X)$, which is equivalent to finding the sequence with the highest joint probability $\arg\max_Y p_\theta(X, Y)$.
In practice, decoding is often carried out by searching the minimal distortion path on a weighted finite state transducer (WFST;~\citealt{mohri1997finite, mohri2002weighted}), obtained by composing WFSTs representing HMM emission, HMM transition, pronunciation model, and language model, respectively.
Moreover, it is common to introduce a weighting factor $\alpha$ to balance the importance of acoustic and language model by re-writing the joint probability as 
\begin{equation}
    p_{\theta,\alpha}(X, Y) \propto p_{AM}(X \mid f_{PM}(Y)) \; p_{LM}(Y)^\alpha,\label{eq:lm_alpha}
\end{equation}
where $\alpha$ is determined by minimizing the edit distance $ED$ on a development set $\{ (\tilde{X}_i, \tilde{Y}_i) \}_{i=1}^{\tilde{N}}$:
\begin{equation}
    \alpha = \arg\min_\alpha \sum_{i=1}^{\tilde{N}}
    ED\left(
        \tilde{Y}_i,\; 
        \arg\max_Y \left(
            \log p_{AM}(\tilde{X} \mid f_{PM}(Y)) + \alpha * \log p_{LM}(Y)
        \right)
    \right).
\end{equation}
In our setting, we use either phoneme error rate (PER) or word error rate (WER) but other choices are possible. 
After $\alpha$ is determined, we decode an utterance $X$ by finding
\begin{equation}
    Y^* = \arg\max_Y p_{\theta,\alpha}(X, Y). \label{eq:lm_dec}
\end{equation}

\subsection{End-to-End Systems for Supervised ASR}
\label{sec:end2end}
More recently, end-to-end approaches that directly parametrize $p_\theta(Y \mid X)$ have gained increasing interest. This line of approaches includes but is not limited to connectionist temporal classification (CTC;~\citealt{graves2006ctc}), recurrent neural network transducer (RNN-T;~\citealt{graves2012sequence}), sequence-to-sequence models (seq2seq;~\citealt{chorowski2015attention}). We focus our discussion on CTC, which is most relevant to this work.

\subsubsection{Training a CTC Model}
As mentioned above, the emission model in the hybrid system can be replaced with a neural network predicting the posterior probability over HMM states for each frame, but it requires a seed HMM-GMM model to derive the forced-alignment.
Instead of maximizing the probability of the derived forced alignment to HMM states, CTC parameterizes the distribution over alignment to text sequences directly, and marginalizes over all possible alignments for the target text sequence during training to compute the posterior directly.
Formally speaking, for an input speech sequence $X$ of $T$ frames, CTC predicts a distribution over $B' = B \cup \{\epsilon\}$ for each input step, where $B$ is the text alphabet and $\epsilon$ is a special blank symbol, representing empty output. 
The probability of an alignment $A = [a_1, \cdots, a_T]$ is defined as $\prod_{t=1}^T p_\theta(a_t \mid X)$.
Each alignment is mapped to a text sequence with a function $g$, which first removes consecutive repeating units and then removes all $\epsilon$. For example, an alignment ``c$\epsilon$aa$\epsilon$abb'' is mapped to a text sequence ``caab''. The posterior probability of a text sequence $Y$ given speech $X$ of length $T$ is therefore defined as:
\begin{equation}
    p_\theta(Y \mid X) = \sum_{A: g(A) = Y, A \in (B')^T} \prod_{t=1}^T p_\theta(a_t \mid X),
\end{equation}
and the marginalization on the right hand side can be computed efficiently with dynamic programming. Training of CTC optimizes the likelihood of the posterior distribution:
\begin{equation}
    p_\theta^* = \arg\max_{p_\theta} \sum_{i=1}^N \log p_\theta(Y_i \mid X_i),
\end{equation}
and decoding is approximated with finding the most probable alignment and mapping that alignment to a text sequence: 
\begin{align*}
    Y^* 
    = g\left(\arg\max_A \prod_{t=1}^T p_\theta(a_t \mid X)\right) 
    = g\left(\prod_{t=1}^T \arg\max_{a_t} p_\theta(a_t \mid X)\right).
\end{align*}

\subsubsection{Decoding CTC with a Language Model Using WFST}
While CTC is motivated by end-to-end modeling of speech recognition, it can also fit nicely into the hybrid framework. The HMM associated with CTC has one state for each letter and a special state for the blank symbol~\citep{zeyer2017ctc}. Each state can transit to any state including itself with the same transition probability. See~\citet{hannun2017sequence} and~\citet{hannun2020differentiable} Figure 3 for an illustration. 
As a result, one can also compose a CTC acoustic model with a letter-PM (mapping words to letters instead of phonemes) and an LM into a single WFST for decoding, or train a CTC model that predicts phonemes to compose with regular PM and LM for decoding.

Likewise, a weighting factor $\alpha$ can be introduced to balance the CTC acoustic model and the language model. By re-arranging \autoref{eq:lm_dec} as
\begin{align*}
    Y^* &= \arg\max_Y p_{AM}(X \mid f_{PM}(Y)) \; p_{LM}(Y)^\alpha 
    = \arg\max_Y p_{\theta}(X, Y) \; p_{LM}(Y)^{\alpha - 1} \\
    &= \arg\max_Y p_{\theta}(Y \mid X) \; p_{LM}(Y)^{\alpha - 1} \\
    &= \arg\max_Y \log p_{\theta}(Y \mid X) + (\alpha - 1) \log p_{LM}(Y),
\end{align*}
we can observe that this is equivalent to the formulation of language model fusion for end-to-end systems that combines posterior and prior probabilities of text for decoding~\citep{chorowski2016towards}.

\subsection{Self-Training for Semi-Supervised ASR}
In the semi-supervised setup, one is provided with unpaired speech $\mathcal{D}_u^s = \{X_j\}_{j=1}^{N_s}$ and text $\mathcal{D}_u^t = \{Y_j\}_{k=1}^{N_t}$ in addition to a labeled dataset $\mathcal{D}_l$. 
Most of the semi-supervised ASR studies focus on how to utilize unpaired speech, because it is straightforward to utilize additional text by training a better language model with both paired and unpaired text for decoding.
Self-training is one of the simplest but most effective approaches to leveraging unpaired speech~\citep{kahn2020st}.
It first trains a seed ASR model using labeled data $\mathcal{D}_l$, and then transcribes unpaired speech $\mathcal{D}_u^s$ into text $\{\hat{Y}_j\}_{j=1}^{N_s}$ using the trained model to generate a pseudo-labeled dataset $\hat{\mathcal{D}}_l = \{ (X_j, \hat{Y}_j \}_{j=1}^{N_s}$. The transcribed text is often called the \emph{pseudo labels}. 
Combining the real and the generated paired data, an ASR model can be trained with any supervised objective~\citep{kahn2020st}, or with separate objectives and weights~\citep{vesely2017semi, manohar2018semi, hsu2020semi}.
This process can also can be repeated for multiple iterations to further improve performance~\citep{xu2020iterative,park2020improved}.

\section{Speech and Text Representations}

In the following, we describe how we build suitable speech and text representations for unsupervised learning.
Good representations are essential to learning a mapping from unlabeled speech audio to unlabeled text in a completely unsupervised fashion.

\subsection{Self-supervised Learning of Speech Audio Representations}
\label{sec:wav2vec2}

In the first step, we learn representations of the speech audio signal using self-supervised learning. 
There has been a lot of recent work in this direction which has shown strong performance in extremely low-labeled data setups across a range of languages~\citep{conneau2020unsupervised} and tasks~\citep{fan2021exploring,pepino2021emotion,wang2021st}. 

Wav2vec 2.0 consists of a convolutional feature encoder $f: \Inp \mapsto \Feat$ that maps a raw audio sequence~$\x$ to latent speech representations $\ze_1, \dots, \ze_T$, which a Transformer $g: \Feat \mapsto \Context$ then turns into context representations $\cc_1, \dots, \cc_T$~\citep{baevski2019vqwav2vec,baevski2019effectiveness}.
Each $\ze_t$ represents about 25ms of audio strided by 20ms and the Transformer architecture follows BERT~\citep{vaswani2017transformer,devlin2018bert}.
During training, latent representations are discretized to $\zq_1, \dots, \zq_T$ with a quantization module $\Feat \mapsto \QFeat$ to represent the targets in the objective.
The quantization module uses a Gumbel softmax to choose entries from $E=2$ codebooks with $V=320$ entries each and the chosen entries are concatenated to obtain $\zq$~\citep{jegou2011ieee,jang2016gumbel,baevski2019vqwav2vec}.

The model is trained to identify the true quantized latent $\zq_t$ for each masked time-step within a set of distractors $Q_t$ sampled from other masked time steps:
$$
-\log \frac{\exp(sim(\cc_t, \zq_t))}{\sum_{\zqt \sim Q_t} \exp(sim(\cc_t, \zqt))}
$$
where $\cc_t$ is the output of the Transformer, and $sim(a, b)$ denotes cosine similarity.

This is augmented by a codebook diversity penalty to encourage the model to use all codebook entries~\citep{dieleman2018challenge}.
The codebook diversity penalty maximizes the entropy of the averaged softmax distribution over the codebook entries for each group $\bar{p}_{g}$ across a batch of utterances: 
$$
    \frac{1}{EV} \sum_{g=1}^{E} - H(\bar{p}_{g}) = \frac{1}{EV} \sum_{g=1}^{E} \sum_{v=1}^{V} \bar{p}_{g,v} \log{\bar{p}_{g,v}}.
$$

In our experiments, we use the publicly available English model pre-trained on 53k hours of \vox{}~\citep{kahn2020librilight} as well as XLSR-53 which was pre-trained on nearly 60k hours of speech audio in 53 languages~\citep{conneau2020unsupervised}. 
Both use Large transformers with 24 blocks, model dimension 1,024, inner-dimension 4,096 and 16 attention heads~\citep{baevski2020wav}.

\subsection{Choosing Audio Representations}
\label{sec:choosing_representations}

Next, we embed the speech audio using self-supervised representations from \wvpp{}.
Before that, we also remove silences from the speech audio using an off-the-shelf unsupervised method. 
One exception is the TIMIT benchmark, where silences are part of the transcription.
Silence removal is important to learning a better mapping between speech audio representations and transcriptions.

\paragraph{Removing Silences.}
Most datasets we use for our experiments have audio data with silences.
However, these parts of the audio do not correspond to any transcription and we therefore remove silences as much as possible.
To remove silences, we apply rVAD, an unsupervised voice activity detection (VAD) model which determines the segments in the audio data corresponding to silences, and we remove these sections~\citep{tan_rvad}.
We ablate this choice in~\autoref{sec:preproc_text}.

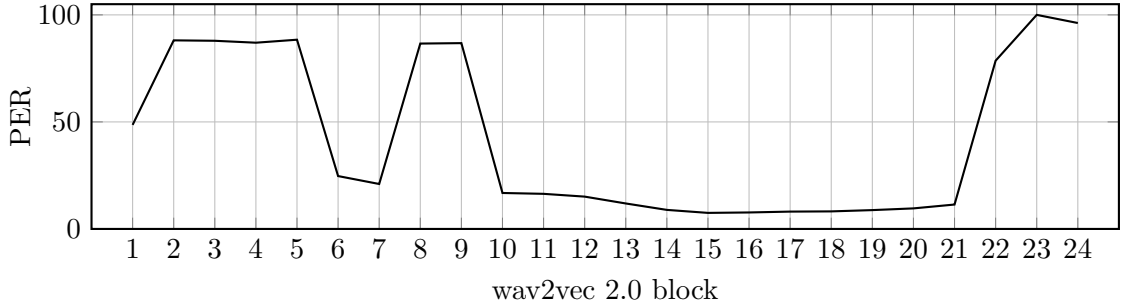
\begin{figure}[t]
\begin{tikzpicture}
\begin{axis}[
width=1.0*\textwidth,
height=.3\textwidth,
legend style={at={(0.97,0.75)},
anchor=east,legend columns=2},
xticklabels from table={\layerdatals}{layer},
xtick=data,
nodes near coords align={vertical},
ymin=0,ymax=105,
xmin=-1,
xmax=24,
ylabel={PER},
ylabel shift = -5 pt,
xlabel={\wvpp{} block},
grid=both,
style=thick
]
\addplot[black] table[x expr=\coordindex,y=en_LS]{\layerdatals};
\end{axis}
\end{tikzpicture}
\caption{Supervised phoneme classification using representations from different \wvpp{} blocks on dev-other of English \libri{}.
Low and high blocks do not provide good features, while as blocks 14-19 do. Block 15 performs best.
}
\label{fig:layers_ls}
\end{figure}

\begin{figure}[t]
\begin{tikzpicture}
\begin{axis}[
width=1.0*\textwidth,
height=.3\textwidth,
legend style={at={(0.84,0.75)},
anchor=east,legend columns=2},
xticklabels from table={\layerdatamls}{layer},
xtick=data,
nodes near coords align={vertical},
ymin=0,ymax=105,
xmin=-1,
xmax=24,
bar width=1pt,
ylabel={PER},
ylabel shift = -5 pt,
xlabel={\wvpp{} block},
grid=both,
style=thick
]
\addplot[black] plot[error bars/.cd, y dir = both, y explicit] table[x expr=\coordindex,y=mean,y error=stdv]{\layerdatamls};
\end{axis}
\end{tikzpicture}
\caption{Supervised phoneme classification on eight languages of the MLS dataset in terms of mean PER and standard deviation for different \wvpp{} blocks to represent the raw audio (cf.~\autoref{fig:layers_ls}). 
We consider English, German, Spanish, French, Italian, Dutch, Polish and Portuguese.
}
\label{fig:layers_mls}
\end{figure}
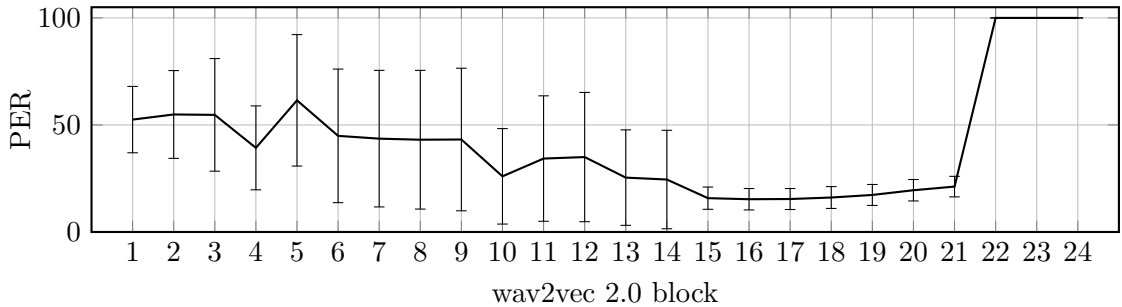

\paragraph{Speech Audio Representations.}
After silence removal, we embed the unlabeled speech audio with \wvpp{} to obtain speech representations.
Specifically, we use the representations of the context Transformer network $\cc_1, \dots, \cc_T$ (\autoref{sec:wav2vec2}).
Each $\cc_t$ represents features computed from the entire utterance using self-attention and centered over the current time-step $t$; there is a 20ms gap between subsequent context representations (\autoref{sec:wav2vec2}).
The context network contains 24 Transformer blocks and we denote the output of block $l$ at time-step $t$ as $\cc_t^l$.

Our goal is to learn a model which can map from audio representations $\cc_t^l$ to phonemes using no supervision.
However, the representations of the uppermost block of \wvpp{} may not be well suited for this task. 
These features are trained to directly predict masked latent representations spanning 25ms of speech audio which is much shorter than the typical duration of a phoneme.
The uppermost block is therefore likely not well suited to learning a good mapping from the current time–step to phonemes. Similarly in natural language understanding, probing Transformer layers of the BERT models showed that blocks behave differently, with earlier blocks encoding syntactic information, while high-level semantic
information appears at higher blocks~\citep{tenney2019bert,jawahar2019does}.

To get a better sense of this, we train supervised phoneme recognizers with a CTC loss on top of the frozen representations of each of the 24 blocks of the English \wvpp{} \wvpplarge{} model pre-trained on \vox{}.
We then evaluate phoneme error rate (PER) with respect to the phonemized transcriptions of \libri{} dev-other.
The classifier takes as input $\cc_t^l$ and contains a single softmax-normalized linear layer mapping to the phoneme inventory.

\autoref{fig:layers_ls} shows that most of the first ten blocks as well as the final blocks provide very poor performance, while blocks 15-19 provide error rates below 9 PER.
Block 15 achieves the best error rate of 7.5 PER.
A similar insight has been used in the concurrent work of~\citet{hsu2020hubert}.
Does this choice of representation generalize across languages? 
To get a sense of this, we measure PER for eight different languages of the MLS dataset (\autoref{sec:datasets}) when inputting representations of the multilingual \wvpp{} XLSR-53 model.
\autoref{fig:layers_mls} confirms that block 15 provides good performance across a range of languages and we will use block 15 as the representation for speech audio in all subsequent experiments.
For brevity we drop the superscript $l$ and refer to block 15 representations simply as $\cc_1, \dots, \cc_T$.

\subsection{Segmenting the Audio Signal}
\label{sec:segment}

Once the speech signal is embedded, we identify segments corresponding to meaningful units that can be mapped to phonemes. 
Segmentation has been shown to be crucial in prior work~\citep{chung2018unsup} since the right boundaries in the input representations make it easier to predict the output sequence.

\paragraph{Identifying Speech Audio Segments.}
There has been a lot of prior work in unsupervised speech segmentation~\citep{kamper2017seg,kamper2017embedded,rasanen2015interspeech,kreuk2020self} but here we simply use a method based on clustering the \wvpp{} speech representations $\cc_1, \dots, \cc_T$.
In a first step, we collect all the speech representations for the unlabeled speech data and perform k-means clustering to identify $K=128$ clusters. 
We use the FAISS library to do fast clustering on GPUs~\citep{johnson2017faiss}.
Next, each $\cc_t$ is labeled with the corresponding cluster ID $\id_t \in \{1, \dots, K\}$ and we introduce speech segment boundaries whenever the cluster ID changes.

\begin{figure}[t]
\centering
\includegraphics[width=1\textwidth]{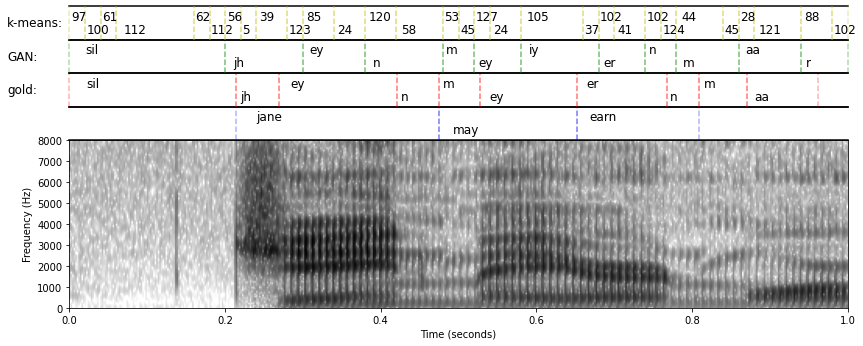} 
\caption{
Example of segmenting the audio signal of the utterance \emph{jane may earn more}.
The top line shows the segmentation by the k-means segmentation method, the second line is the segmentation of Viterbi decoding with the GAN, and the third line shows the gold segmentation of human annotators from the TIMIT dataset.
At the bottom, we show the corresponding spectogram, although, the input to our method is raw audio. 
The utterance has TIMIT ID MGLB0\_SX4.
}
\label{fig:segment_ex}
\end{figure}

\paragraph{Building Segment Representations.}
Once the speech audio representations are segmented, we compute a  512-dimensional PCA over all speech representations output by \wvpp{} for the training set. 
Next, we mean-pool the PCA representations for a particular segment to obtain an average representation of the segment. 
The PCA retains only the most important features and we found this to be effective.
Segment boundaries are noisy due to the lack of supervision and we therefore found it useful to also mean-pool pairs of adjacent segment representations to increase robustness.
This results in sequences of speech segment representation $\s = \sss_1, \dots, \sss_T, \s \sim \Seg$ for a given utterance.

\autoref{fig:segment_ex} illustrates how the k-means segmentation strategy results in very granular units compared to the gold segments corresponding to phonemes. 
Based on the k-means units, the unsupervised model can then recover segments that correspond very closely to phonemic units identified by humans.

\insertTIMITBoundary

\autoref{tab:boundary} shows that k-means clustering results in very high precision but low recall when recovering gold phoneme boundaries on TIMIT. 
The Viterbi outputs of our model (\wvz{}) result in more balanced, and better, accuracy because neighboring segments with the same predicted label are combined into a larger segment.

\subsection{Pre-processing the Text Data}
\label{sec:preproc_text}

Similar to how we segment the unlabeled speech audio data into suitable units for unsupervised learning, we do the same for the unlabeled text data.
We apply two pre-processing steps to the text data: phonemization and silence token insertion.

\paragraph{Phonemization.}
Phonemes characterize the different sounds which distinguish words from each other, e.g., for the word \emph{cat} there are three phonemes corresponding to the three distinct sounds in the pronunciation of the word: /K/, /AE/, /T/.
We phonemize the text data because we found it easier to learn a mapping between speech audio and the different sounds of a word rather than between audio and words or letters.
Phonemization converts a sequence of words $Y$ into a sequence of phonemes $P = [p_1, \cdots, p_M]$, where $p_m \in O$ and $O$ is the phoneme inventory.
We use off-the-shelf tools for this step which we detail in \autoref{sec:setup_phonemization}.

\paragraph{Silence token insertion.}
The unlabeled speech audio data is pre-processed by applying unsupervised silence removal.
However, this process is not always accurate and many silences in the speech audio remain.
To deal with this, we enable the unsupervised model to label some segments with a phonemic silence token (SIL;~\autoref{sec:architecture}).
However, the phonemized unlabeled text data does not contain any silence tokens and this may pose difficulties for adversarial learning (\autoref{sec:unsupervised_learning}).
We remedy this by inserting silence markers into the phonemized unlabeled text data.

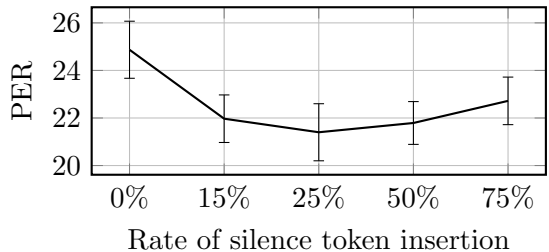
\begin{figure}[t]
\centering
\begin{minipage}{0.44\textwidth}
\vspace{-0.9cm}
\insertSilence
\end{minipage}
\hfill
\begin{minipage}{0.5\textwidth}
\centering
\begin{tikzpicture}
\begin{axis}[
width=1*\textwidth,
height=.5\textwidth,
xticklabels from table={\silencetokdata}{perc},
xtick=data,
nodes near coords align={vertical},
ylabel={PER},
xlabel={Rate of silence token insertion},
grid=both,
style=thick,
]
\addplot[black] plot[error bars/.cd, y dir = both, y explicit] table[x expr=\coordindex,y=mean,y error=stdv]{\silencetokdata};
\end{axis}
\end{tikzpicture}
\end{minipage}
\caption{Unsupervised performance when augmenting the unlabeled text data with silence tokens.
We add silence tokens to the unlabeled text to better resemble the speech audio which does contain silences.
Silence tokens surrounding sentences and removing silences from the audio results in better performance (left), and we show different rates of silence token insertion in the unlabeled text data (right).
We report mean PER and standard deviation over 20 random seeds of unsupervised training on \libri{} dev-other.
}
\label{fig:ablation_silence}
\end{figure}

First, we add a SIL token to the beginning and the end of all phonemized unlabeled text sentences. 
\autoref{fig:ablation_silence} (left) shows that this improves accuracy.
The same table shows the detrimental effect of not removing silences from the audio data (\autoref{sec:choosing_representations}).
Second, we randomly insert SIL between words, or groups of phonemes corresponding to words.
\autoref{fig:ablation_silence} (right) shows that inserting the silence token at a rate of 0.25 yields the best end accuracy.

\section{Unsupervised Learning}
\label{sec:unsupervised_learning}

We use adversarial training to train an unsupervised speech recognition model using the representations of the unlabeled speech audio data and the unlabeled phonemized text data ~\citep{liu2018completely,chen2019completely}.
In the following, we detail the model architecture, the training objective as well as the unsupervised cross-validation metric we developed.

\subsection{Model Architecture}
\label{sec:architecture}

Generative adversarial networks (GAN;~\citealt{goodfellow2014generative}) train a generator network \gen{} and a discriminator/critic network \dis{} where the generator produces samples which are then judged by the discriminator.
The discriminator is trained to classify whether samples are from the generator or from the real data distribution. 
The objective of the generator is to produce samples that are indistinguishable by the discriminator.

Concretely, \gen{} takes as input a sequence of $T$ segment representations $\s = \left[ \sss_1, \dots, \sss_T \right]$ (\autoref{sec:segment}) which are then mapped to a sequence of $M$ phonemes $\genm(\s) = \left[\pp_1, \dots, \pp_M \right]$.
The generator predicts a distribution over the phoneme set $O$ for each segment and outputs the phoneme with the highest probability.
If the argmax prediction of consecutive segments result in the same phoneme, then we sample one of these segments, therefore $M \leq T$.

The phoneme set $O$ includes a silence label SIL to enable labeling silences in the speech audio as such.
Without a silence label, we noticed that the model was repurposing a particular phoneme to label silences which resulted in much lower performance since it interfered with subsequent language model decoding. %
In the backward pass, we back-propagate through segments sampled at the generator output.
We do not modify the segment representations $\s$ during unsupervised training.
The generator is parameterized as a single layer convolutional neural network (CNN;~\autoref{sec:train_details}).

\begin{figure}[t]
\centering
\includegraphics[width=0.65\textwidth]{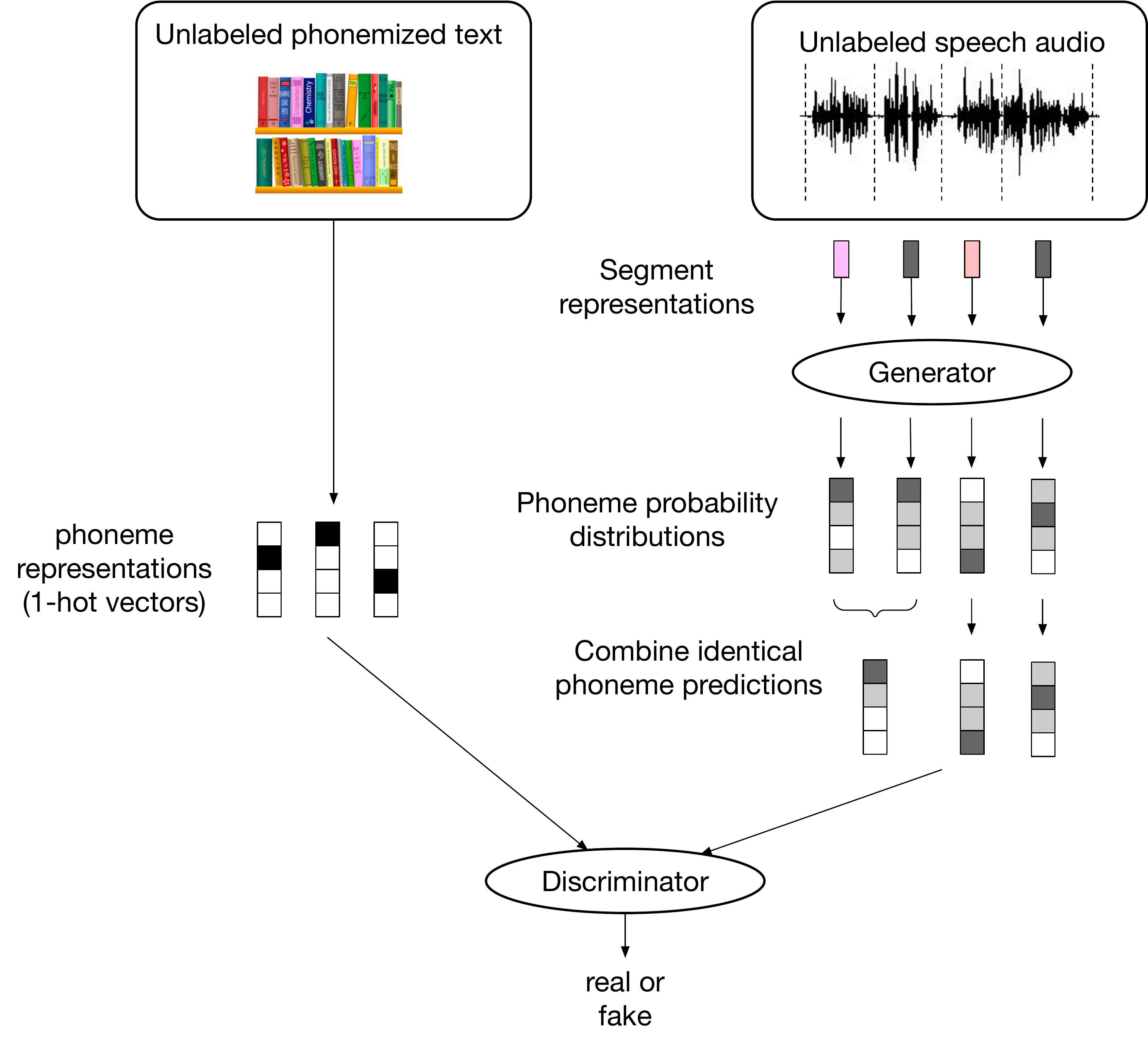} 
\caption{
Illustration of how generator outputs and real phonemized text are turned into inputs to the discriminator. Real text is represented as a sequence of 1-hot vectors and generator outputs for different segment representations are collapsed if consecutive segments result in the same argmax prediction.
}
\label{fig:discriminator}
\end{figure}

The discriminator takes as input either a sequence $\pr \sim \Phonr$ of one-hot vectors denoting phonemized text from the real data distribution $\Phonr$ or a sequence of output distributions from the generator $\genm(\s)$.
Each input vector has $|O|$ dimensions to represent the distribution over phonemes for each segment (see \autoref{fig:discriminator} for an illustration).
The discriminator is also a CNN which outputs a probability indicating how likely the sample is to be from the data distribution (\autoref{sec:train_details}).

\subsection{Objective}
\label{sec:obj]}

In our setup we use the original GAN objective with a gradient penalty~\citep{goodfellow2014generative,arjovsky2017wasserstein}, a segment smoothness penalty and a phoneme diversity penalty:
\begin{equation}
    \min_{\genm}~\max_{\dism}~~ 
    \E_{\pr \sim \Phonr} \left[ \log \dism(\pr) \right] +
    \E_{\s \sim \Seg} \left[ \log \left(1 - \dism (\genm(\s)) \right) \right]
    - \lambda \mathcal{L}_{gp}
    + \gamma \mathcal{L}_{sp}
    + \eta \mathcal{L}_{pd}    
\end{equation}
where $\pr \sim \Phonr$ is phonemized unlabeled text, $\genm(\s)$ is the transcription output by the generator of input segment representations $\s$ for some unlabeled speech audio.
The first term trains the discriminator to assign high probability to real transcriptions, the second term encourages the discriminator to assign low probability to generator outputs, $\mathcal{L}_{gp}$ is a gradient penalty, $\mathcal{L}_{sp}$ is a smoothness penalty and $\mathcal{L}_{pd}$ is a phoneme diversity loss which we detail next.
During training we alternate updates for the discriminator and the generator. 
We also alternate batches of predicted transcriptions from the generator and phonemized unlabeled text.

\paragraph{Gradient penalty.}
To stabilize training, we penalize the gradient norm of the discriminator with respect to the input~\citep{gulrajani2017improved}.
The penalty is computed for random samples $\prand \sim \mathcal{\tilde{P}}$ which are a linear combination of the activations of pairs of real and fake samples.\footnote{We simply shorten longer sequence if the lengths differ.}
\begin{equation}
\mathcal{L}_{gp} = \E_{\prand \sim \mathcal{\tilde{P}}} \left[ \left( \| \nabla \dism (\prand) \| - 1 \right)^2 \right]
\end{equation}

\paragraph{Segment smoothness penalty.}
The k-means segmentation of the speech audio is more granular than a typical phonemized transcription and neighboring representations are highly correlated.
We therefore found it useful to add a penalty which encourages the generator to produce similar outputs for adjacent segments:
\begin{equation}
\mathcal{L}_{sp} = \sum_{(\pp_t, \pp_{t+1}) \in \genm(\s)} \| \pp_t - \pp_{t+1} \| ^2
\end{equation}
where $\pp_t \in \mathbb{R}^{|O|}$.

\paragraph{Phoneme diversity loss.}
We also found it helpful to penalize low usage of the phoneme vocabulary by the generator network on the batch level. 
In particular, we maximize the entropy of the averaged softmax distribution $H_\genm(\genm(\s))$ of the generator over the phoneme vocabulary across a batch $B$ of utterances:
\begin{equation}
\mathcal{L}_{pd} = \frac{1}{|B|} \sum_{\s \in B} -H_\genm(\genm(\s))
\end{equation}

\subsection{Unsupervised Cross-Validation Metric}
\label{sec:metrics}

Our goal is to build speech recognition models without any supervision. 
To this end, we developed the following cross-validation metric which does not require labeled data.
We use the metric for early stopping, selecting a random seed, and hyper-parameter selection ($\lambda$, $\gamma$, $\eta$).

We consider two quantities in our metric: \emph{LM negative log-likelihood (NLL)} and \emph{vocabulary usage}.
$NLL_{LM}$ serves as an indicator of fluency for a given transcription and it is measured with a language model $p_{LM}$ trained on phonemized text data (\autoref{sec:preproc_text}). 
Vocabulary usage is the proportion of the phoneme vocabulary being output by the model via Viterbi decoding.
Measuring vocabulary usage identifies degenerate models which output fluent but trivial transcriptions.

We denote Viterbi phoneme transcriptions for a given generator configuration \gen{} and unlabeled speech audio $\{\x_j\}^{N_s}_{j=1}$ as $\mathcal{\p} = \{\p_j\}^{N_s}_{j=1}$.
$NLL_{LM}$ is measured in the standard way over the phonemized transcriptions: $NLL_{LM}(\mathcal{\p}) = \frac{1}{N_s} \sum^{N_s}_{j=1} NLL_{LM}(\p_j)$ where $NLL_{LM}(\p) = -\frac{1}{M} \sum^M_{t=1} \log p_{LM}(p_t)$ using $p_{LM}(p_t)$ as shorthand for $p_{LM}(p_t|p_{t-1}, \dots, p_1)$.\footnote{We remove SIL labels from $\mathcal{\p}$ when computing the NLL because SIL is not used in $p_{LM}$.}
On the other hand, we use $U(\mathcal{\p}) = \frac{1}{|O|} {\sum_{o \in O} [o \in \mathcal{\p}]} \in [0,1]$ to denote the vocabulary usage of $\mathcal{\p}$.

In the first step, we generate phoneme transcriptions for different training checkpoints or hyper-parameter settings and denote the transcriptions of the configuration with the lowest vocabulary-usage adjusted NLL as $\mathcal{\pbest} = \arg\min_{\mathcal{\p}} NLL_{LM}(\mathcal{\p}) - \log U(\mathcal{P})$.\footnote{In practice, we used language model perplexity which is equivalent to NLL after taking the log.} 
Next, we discard model configurations which do not satisfy the following, using $\mathcal{\pbest}$ as the anchor:
\begin{equation}
NLL_{LM}(\mathcal{\p}) < NLL_{LM}(\mathcal{\pbest}) + \log \left( \frac{U(\mathcal{\p})}{U(\mathcal{\pbest})} \right) + \log 1.2
\end{equation}
The second term on the right hand side introduces a margin over the NLL of the anchor transcription $NLL_{LM}(\mathcal{\pbest})$ based on the vocabulary usage of $\mathcal{\p}$ and $\mathcal{\pbest}$: 
If $U(\mathcal{\pbest})$ is much lower compared to $U(\mathcal{\p})$, then we allow model configurations which produce transcriptions with higher NLL compared to $\mathcal{\pbest}$.
However, if $U(\mathcal{\pbest})$ is a lot higher than $U(\mathcal{\p})$, then the model configuration will not satisfy the constraint. 
The $\log 1.2$ factor serves as another margin allowing checkpoints with slightly worse vocabulary-usage adjusted NLL to be included.

In the final step, we take into account the length of the transcriptions: 
out of the configurations $\mathcal{\p}'$ which satisfy the above constraint, we select the one which has the highest sum of log probability without normalizing the length:
\begin{equation}
\mathcal{\p}^* = \arg\max_{\mathcal{\p}'} \sum^{N_s}_{j=1} \sum^M_{t=1} \log p_{LM}(\pp^j_t), M = |P^j|, \p^j = [\pp^j_1,\dots,\pp^j_M]
\end{equation}
This selects model configurations which produce phoneme sequences that score high under the language model but are not too long.

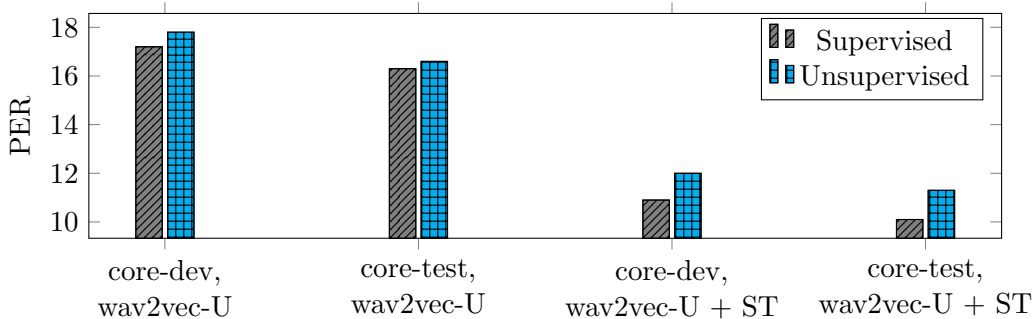
\begin{figure*}[t]
\centering
\begin{tikzpicture}
\begin{axis}[
ybar,
width=0.9*\textwidth,
height=.3\textwidth,
xticklabels from table={\metricdata}{setting},
xticklabel style={text width=3.2cm, align=center},
xtick=data,
nodes near coords align={vertical},
ylabel={PER},
]
\addplot[black,fill=gray,postaction={pattern=north east lines}] table[x expr=\coordindex,y=lbld]{\metricdata};
\addplot[black,fill=cyan,postaction={pattern=grid}] table[x expr=\coordindex,y=unlblbd]{\metricdata};
\legend{Supervised, Unsupervised}
\end{axis}
\end{tikzpicture}
\caption{Effectiveness of the unsupervised cross-validation metric for model development compared to using a labeled development set (Supervised). 
We report PER on TIMIT core-dev/test (\autoref{sec:datasets}) for the GAN (\wvz{}) and with self-training (\wvz{} + ST).
}
\label{fig:metric}
\end{figure*}

How effective is this metric?
To get a sense of this, we compare the performance of cross-validation with a labeled development to cross-validation with our unsupervised metric on the TIMIT benchmark. 
We cross-validate GAN hyper-parameters, different checkpoints for early stopping, language model decoding hyper-parameters (\autoref{sec:decoding}) and HMM decoding hyper-parameters for self-training (\autoref{sec:setup_selftrain}).
\autoref{fig:metric} shows that the unsupervised metric has only between 0.3-1.2 higher PER compared to using a labeled development set. 
This enables model development of unsupervised speech recognition systems without labeled data at only a small drop in accuracy compared to the ideal setting where labeled development data is available.

\section{Experimental Setup}

\subsection{Datasets}
\label{sec:datasets}

We consider several corpora and a variety of languages to evaluate our approach.
TIMIT is a small English dataset on which previous unsupervised speech recognitition work was conducted. 
\libri{} is a standard benchmark with nearly 1,000 hours of labeled English speech audio and MLS is a multilingual benchmark with eight European languages.
In addition, we also consider non-European languages from the ALFFA and CommonVoice corpora.

\paragraph{TIMIT.}
This dataset contains about five hours of audio recordings with time-aligned phonetic transcripts~\citep{garofolo1993timit}. 
To compare to prior work, we consider two setups: the \emph{matched} setting uses text and speech from the same set of utterances to train the model while the \emph{unmatched} setting ensures that the unlabeled text data does not contain the transcriptions of the audio data.
For the matched setup, we follow the standard train/dev/test split of TIMIT as done in~\citet{yeh2018unsupervised}. 
This is 3,696/400/192 train/dev/test utterances which contains only SX (compact) and SI (diverse) sentences.
For the unmatched setting, we follow~\citet{chen2019completely} by training on 3,000 speech utterances and 1,000 transcriptions from the training portion of the complete dataset split. 
We use the remaining 620 training utterances for validation, and test on 1,680 sentences for testing.
The complete dataset split contains 4,620 training and 1,680 testing utterances, with additional SA (dialect) sentences.

\paragraph{\libri{} and \vox{}.}
The \libri{} corpus contains 960 hours of transcribed speech audio (\librisz{}) for training. 
The data is based on read English audio books.
For unsupervised training, we only use the speech audio data but not the transcriptions.
We use the official \libri{} language modeling data as unlabeled text data with the \vox{} data removed~\citep{synnaeve2020end}.\footnote{\url{https://github.com/flashlight/wav2letter/tree/master/recipes/sota/2019\#non-overlap-lm-corpus-librispeech-official-lm-corpus-excluded-the-data-from-librivox}}
This is a large text corpus of 635m words but we show that much smaller amounts of text and speech data still result in the same performance (\autoref{sec:exp_datasize}).
We evaluate on the standard dev-other/clean and test-clean/other sets.
For development, we compute the unsupervised metric (\autoref{sec:metrics}) on dev-other.
We also experiment with the audio data from \vox{} (\voxsz{}) for which we follow the pre-processing of~\citet{kahn2020librilight} resulting in 53.2k hours of speech audio.

\paragraph{Multilingual LibriSpeech (MLS).} 
The Multilingual Librispeech dataset~\citep{pratap2020mls} is a large corpus of read audiobooks from Librivox in eight languages and we experiment with the following six languages: \textit{Dutch (du), French (fr), German (de), Italian (it), Portuguese (pt), Spanish (es)}. 
The latest version of this corpus contains around 50k hours including 44k hours in English. 
However, for unsupervised learning we only use 100 hours of speech audio for each language.
As unlabeled text data, we use the LM data provided by MLS.

\paragraph{ALFFA.}
We experiment with the Swahili data from the ALFFA project~\citep{gelas2012alffa,abate2005amharic,tachbelie2014alffa} which is read speech.
There are 9.2 hours of speech audio training data and we use the language modeling data provided by ALFFA as unlabeled text data as well as newscrawl 2008-2014, 2018-2020.\footnote{\url{http://data.statmt.org/news-crawl/sw}}

\paragraph{CommonVoice.}
This is a multilingual corpus of read speech for 38 languages~\citep{ardila2019common}.
We focus on two low-resource languages Kyrgyz (ky) and Tatar (tt) and use 1.8 hours and 4.6 hours of speech audio, respectively.
As unlabeled text data for Kyrgyz we use the Kyrgyz community corpus 2017,\footnote{\url{https://corpora.uni-leipzig.de?corpusId=kir\_community_2017}} and newscrawl 2008-2014 and 2018-2020.\footnote{\url{http://data.statmt.org/news-crawl/ky}}
For Tatar we use the Tatar community corpus 2017,\footnote{\url{https://corpora.uni-leipzig.de?corpusId=tat\_community_2017}} and newscrawl 2005-2011.\footnote{\url{https://corpora.uni-leipzig.de/en?corpusId=tat_news_2005-2011}}
We evaluate on the dev and test split of~\citet{rivire2020unsupervised}.

\subsection{Phonemization}
\label{sec:setup_phonemization}

TIMIT provides time-aligned phonetic transcriptions annotated with an inventory of 60 phones adapted from the ARPAbet system, which treats silence as a phoneme. 
In addition, it includes a mapping from the 60 phoneme inventory to 48- and 39-phoneme inventories. 
Phoneme error rates are typically computed on the 39-phoneme inventory~\citep{povey2011kaldi}, which we map the phonetic transcripts to for training.

For Librispeech, we use the G2P phonemizer~\citep{g2pE2019} which uses the CMU dictionary to look up English word pronunciations, falling back to a neural network trained to output a phoneme sequence given a word. 
We phonemize the Librispeech LM corpus, with Librispeech and Librivox text data removed~\citep{synnaeve2020end}.
We convert the full phoneme set to a reduced set containing 39 phonemes by removing the numerical stress markers from the vowels.

For other corpora, including English in the MLS dataset, we use Phonemizer which supports a large number of various languages, but is less accurate than G2P for English.\footnote{\url{https://github.com/bootphon/phonemizer}}
We disable the language-switching labels and prune phonemes that appear fewer than 1000 times in the text corpus.

\subsection{Unsupervised Training Details}
\label{sec:train_details}

Models are implemented in fairseq~\citep{ott2019fairseq}.
The generator and discriminator are optimized with Adam \citep{kingma2015adam} using $\beta_1=0.5$ and $\beta_2=0.98$. 
The discriminator has a weight decay of $1e-4$ while the generator does not use weight decay. The discriminator is trained with a learning rate of $1e-5$ and the generator with $1e-4$, which are held constant throughout the training. We train for a total of 150k steps, during which we alternate optimizing the discriminator and the generator (so each are updated 75k times in total). Each training step is performed using a batch of 160 randomly chosen samples from the unlabeled audio data and 160 randomly chosen text samples from the unlabeled text data.
Training takes about 12 hours on a single V100 GPU.

The discriminator is composed of three causal convolution blocks with a hidden size of $384$ and a kernel size of $6$, resulting in a receptive field size of 16 segments. 
The input into the discriminator is an $|O|$ dimensional vector representing the probability distribution over the phoneme vocabulary, and the output is a single logit for each time-step, indicating how likely the sample is to be from the data distribution.
The first layer serves as embedding for the $|O|$ phonemes.

The generator is a single non-causal convolution with kernel size 4. The input to the generator are the segment representations $S$ of dimension 512 and the output is an $|O|$ dimensional vector. 
The generator contains about 90k parameters and we do not backpropagate to the segment representations.
We combine subsequent generator predictions prior to feeding them into the discriminator as described in~\autoref{sec:architecture} and apply a softmax normalization. During training, we use dropout with $p=0.1$ to the input of the generator~\citep{srivastava2014dropout}.

For each language, we tune the following hyper-parameters using the unsupervised cross-validation metric (\autoref{sec:metrics}): the gradient penalty weight $\lambda$ is selected from the range $[1.5,2.0]$, the smoothness penalty weight $\gamma$ from $[0.5,0.75]$, the phoneme diversity loss weight $\eta$ from $[2,4]$, and we train 5 seeds for each configuration for a total of 40 models.

\subsection{Decoding}
\label{sec:decoding}

We wish to decode the output of either phoneme-based models, resulting from unsupervised training, or letter-based models, resulting from subsequent self-training (\autoref{sec:setup_selftrain}) using a language model.

To do so we build WFSTs (\autoref{sec:background_wfst}) using PyKaldi~\citep{can2018pykaldi}, a Python port of Kaldi~\citep{povey2011kaldi}.
The WFST takes as input the model emissions and if we decode to words, then we use the same phonemizer with which we pre-processed the unlabeled text data to build a mapping between phonemes to words.
The WFST is composed with a 4-gram language model~\citep{heafield-2011-kenlm} pruned to keep only 4-grams occurring more than 3 times. 
We add self-loops that mimic CTC behavior~\citep{zhang2020faster} where blank symbols (silence for the GAN or actual blank symbol for letter models) are mapped to epsilons and consecutive predictions of the same symbol are collapsed.

During decoding, we average the predicted phoneme distributions for segments which have the same argmax prediction. 
We provide acoustic scale as a parameter to the Kaldi decoder and we also add a scalar $\nu$ to the blank token emission (silence for GAN models and blank for others). 
We tune the optimal weights for these two parameters by minimizing a quantity that measures fluency of the output as well as faithfulness to the model output. 
In particular we minimize the following quantity on an unlabeled development set - assuming a phoneme-based model which we wish to decode to words:
\begin{equation}
\sum_{j=1}^{N_s} H_{LM}(\bar{\p}_j) \times \max\left(ED(\bar{\p_j}, \p_j), \mu \right)    
\end{equation}
where $\{ \p_j \}^{N_s}_{j=1}$ are the Viterbi model outputs, $\{ \bar{\p}_j \}^{N_s}_{j=1}$ are the word-based outputs of the WFST converted to phonemes (or simply phoneme-based outputs if decoded to phonemes), $H_{LM}(\p_j)$ is the entropy of a language model, $ED$ is an edit distance such as PER for phonemes or character error rate for letter-based model models, and $\mu = 0.03$. 
In practice, trivial solutions may achieve very low entropy. We counteract this by replacing $H_{LM}(\bar{\p}_j)$ by the average entropy of the language model training data if $H_{LM}(\bar{\p}_j)$ is lower than the entropy of the training data.
We tune acoustic scale in the interval $[0, 8]$, and $\nu$ in $[-3,8]$.

For \libri{} experiments, we also decode with a Transformer language model~\citep{baevski2018adaptive} trained on the \libri{} LM corpus using the beam search decoder of~\cite{pratap2019w2l}.
The Transformer LM is identical to~\citet{synnaeve2020end} and contains 20 blocks, model dimension 1,280, inner dimension 6,144 and 16 attention heads. 
We tune acoustic scale with a beam of 50 and test performance is measured with beam 500.

\subsection{Self-Training}
\label{sec:setup_selftrain}

For self-training, we perform two iterations: 
first, we pseudo-label the training data with the unsupervised GAN model and train an HMM on the pseudo-labels (\autoref{sec:hmm}).
Second, we relabel the training data with the HMM and then fine-tune the original \wvpp{} model using the HMM pseudo-labels with a CTC loss~(\autoref{sec:end2end}).
HMM models use phonemes as output, while as \wvpp{} models use letters. 
Both are decoded using WFST decoders into words.
\wvpp{} self-training for \libri{} uses the same fine-tuning parameters as the original \wvpp{} model fine-tuned on 100 hours of \libri{} data, but we reduce masking probability to 0.25, reduce the batch size to 800k frames and train on 8 GPUs for 80k updates.\footnote{\url{https://github.com/pytorch/fairseq/blob/master/examples/wav2vec/config/finetuning/vox\_100h.yaml}}
We use the last checkpoint instead of early stopping.
For TIMIT self-training, we use the one hour fine-tuning parameters of the original \wvpp{} model.\footnote{\url{https://github.com/pytorch/fairseq/blob/master/examples/wav2vec/config/finetuning/vox\_1h.yaml}}
This performs 13k updates on 4 GPUs.

\insertlibrispeech

\section{Results}

We first evaluate our approach on the \libri{} benchmark to compare to competitive systems trained on 960 hours of labeled data (\autoref{sec:exp_libri}).
This is followed by a comparison to prior unsupervised work on the small-scale TIMIT benchmark (\autoref{sec:exp_timit}). 
To get a sense of how well our approach works on non-English languages we evaluate on several European languages of the MLS benchmark (\autoref{sec:exp_mls}) as well as non-European low-resource languages (\autoref{sec:exp_lowres}).
We evaluate the effectiveness of different self-training strategies (\autoref{sec:exp_selftrain}) and show that good performance can be achieved with very little unlabeled data (\autoref{sec:exp_datasize}).

\subsection{Comparison to Supervised Speech Recognition on Librispeech}
\label{sec:exp_libri}

We first test our approach on \libri{} to get a sense of how viable unsupervised speech recognition can be compared to the best supervised systems trained on a large amount of labeled data.
\libri{} is a standard benchmark in the speech recognition community which provides about 960 hours of transcribed read audiobooks.
We use the language modeling data of \libri{} as unlabeled text data for unsupervised training.\footnote{
Below we show that much less unlabeled text and speech audio are sufficient to reach a similar level of performance (\autoref{sec:exp_datasize}).}
We experiment with the frozen representations of 
a \wvpp{} \wvpplarge{} model trained on the 53.2k hours of \vox{} (\voxsz{}) which we denote as \wvz{} \wvpplarge{} and 
we also consider self-training (\autoref{sec:setup_selftrain}). 

\wvz{} \wvpplarge{} with self-training (\wvz{} + ST) and a Transformer language model achieves WER 5.9 on test-other, the noisy test set.
This shows that unsupervised speech recognition can perform remarkably well compared to the best supervised systems of the recent past on this much studied benchmark.
Also, self-training is effective even when the teacher model is unsupervised as per the improvement over GAN training (\wvz{}).
Interestingly, self-training on just \libri{}, or 960 hours of unlabeled speech audio, achieves already very good performance of WER 6.4 on dev-other compared to self-training on all of \vox{} (53.2k hours) which compares at 6.0 WER.
We note that the number of parameters trained during adversarial training is very small:
the generator contains only about 90k parameters for a single temporal convolution mapping to the phoneme set from frozen \wvpp{} representations.

\insertTIMIT

\subsection{Comparison to Prior Unsupervised Work}
\label{sec:exp_timit}

Prior work on unsupervised speech recognition focused on the TIMIT benchmark.
In order to perform a direct comparison to these approaches, we report results on this benchmark as well.
We consider two setups to compare to previous work:
in the matched setting, the unlabeled text data is simply the transcriptions of the unlabeled audio data but unpaired.
In the unmatched setup, the unlabeled text data does not contain the transcriptions for the audio data which is a more realistic setting.

We measure performance on the standard Kaldi dev and test sets (core-dev/core-test) as well as a slightly larger version of the test set (all-test) to be able to compare to \citet{liu2018completely} and \citet{chen2019completely}. Further details of the two setups can be found in \autoref{sec:datasets}.
We report performance for \wvz{} with a 4-gram language model trained on the language modeling data of TIMIT and we also consider self-training (\wvz{} + ST). 

\autoref{tab:timit} shows that \wvz{} outperforms prior unsupervised work in both the matched and unmatched settings, reducing PER on all-test in the matched setup by 57\% relative compared to \citet{chen2019completely}.
Our method has lower performance than the best supervised methods but it performs still very well at PER 12 on core-test in the matched setup compared to PER 8.3 for the state of the art~\citep{baevski2020wav}.

\subsection{Performance on non-English languages}
\label{sec:exp_mls}

\insertMLS

To get a sense of how well the method works on non-English data, we experiment on six languages of the multilingual Librispeech corpus (MLS;~\citealt{pratap2020mls}).
As baseline we consider the supervised systems of~\citet{pratap2020mls} trained on between 2k and 161 hours of labeled data, depending on the language.
For adversarial learning we use 100 hours of unlabeled audio data from MLS for every language as well as the MLS language modeling data.
As input to \wvz{} we use the representations from XLSR-53~\citep{conneau2020unsupervised}, a \wvpp{} model pre-trained on 53 languages.
\autoref{tab:mls} shows that \wvz{} generalizes across a range of languages. 
Performance is lower than supervised systems but it shows the viability for other languages.

\subsection{Application to Low-resource Languages}
\label{sec:exp_lowres}

\begin{table}[t]
\begin{minipage}{0.46\textwidth}
\vspace{-0.1cm}
\centering
\insertLowresourceCV
\end{minipage}
\hfill
\begin{minipage}{0.46\textwidth}
\centering
\insertLowresourceALFFA
\end{minipage}
\end{table}

Experiments so far focused on European languages, for most of which relatively large amounts of labeled data exist.
Next, we turn to three low-resource languages, Swahili, Kyrgyz, and Tatar.
Swahili is an African language, Kyrgyz and Tatar are Turkic languages with only about 4.3m and 5.2m speakers, respectively.\footnote{\url{https://en.wikipedia.org/wiki/{Kyrgyz,Tatar}_language}}
We use between 1.8 hours (Kyrgyz) and 9.2 hours of unlabeled audio (Swahili), see~\autoref{sec:datasets}. To compare to prior work, we measure WER for Swahili and PER for Kyrgyz and Tatar.
For Tatar and Kyrgyz we opted to use a reduced self-training regime for faster experimental turn-around where we only perform HMM self-training and we expect better performance with the full self-training setup (\autoref{sec:exp_selftrain}).
\autoref{tab:lowres_cv} and \autoref{tab:lowres_alffa} show that \wvz{} achieves good performance on these low-resource languages compared to previous work that utilized labeled data.
We note that for Tatar and Kyrgyz we use a much smaller amount of speech audio than prior work: compared to XLSR-53 we use 1.8h unlabeled data vs 17h of labeled data for Kyrgyz and 4.6h vs. 17h for Tatar.

\subsection{Self-training Strategies}
\label{sec:exp_selftrain}

\insertTIMITselftrain

The self-training strategy we use is as follows: once the GAN is trained, we use it together with a language model to pseudo-label the unlabeled audio, then we train an HMM on the labels and repeat pseudo-labeling with the HMM in order to fine-tune the \wvpp{} model whose representation were originally fed to the GAN. 
Finally, we use the fine-tuned \wvpp{} model to decode the test set.

\autoref{tab:timit_selftrain} shows that fine-tuning with an HMM (\wvz{} + HMM) leads to substantial improvements, however, a second iteration of HMM self-training leads to much smaller additional gains (\wvz{} + HMM + HMM).
Using the HMM to re-segment the speech audio followed by repeated GAN training (\wvz{} + HMM resegment + GAN), similar to~\citet{chen2019completely}, does not improve performance over just HMM self-training.

Another option is to directly fine-tune \wvpp{} on the labels assigned by the GAN model (\wvz{} + fine-tune) and this performs very well.
However, another round of self-training based on a fine-tuned \wvpp{} model does not improve performance (\wvpp{} + fine-tune + fine-tune).
We believe that this is due to overfitting since the fine-tuned model has over 300m parameters.
This is in line with recent observations about overfitting in self-training for speech recognition~\citep{likhomanenko2021slimipl}.

The HMM is less likely to overfit in the way the \wvpplarge{} \wvpp{} model does. 
We therefore found it effective to perform a single round of HMM self-training followed by \wvpp{} fine-tuning (\wvz{} + HMM + fine-tune). 
After two rounds of self-training, we do not require a language model anymore for this benchmark which is likely because the language model has been distilled into the model to a large degree.

\subsection{Amount of Unlabeled Data Needed}
\label{sec:exp_datasize}

For the experiments on \libri{} we used large amounts of unlabeled speech audio and text data for adversarial learning (960 hours of unlabeled speech audio and nearly 31m sentences of text data).
For TIMIT we used much less data, only about 3.15h of speech audio and 140k phonemes of text data for the matched setup.
Next, we perform controlled experiments on \libri{} to get a sense of how much data is sufficient to achieve good performance.

\begin{figure}[t]
\centering
\begin{subfigure}[t]{.52\textwidth}
\centering
\begin{tikzpicture}
\begin{axis}[
width=1*\textwidth,
height=.5\textwidth,
xticklabels from table={\audiodata}{nhours},
xticklabel style={font=\small},
xtick=data,
nodes near coords align={vertical},
ymin=19,ymax=26,
ylabel={PER},
xlabel={Amount of audio data},
grid=both,
style=thick,
]
\addplot[black] plot[error bars/.cd, y dir = both, y explicit] table[x expr=\coordindex,y=mean,y error=stdv]{\audiodata};
\end{axis}
\end{tikzpicture}
\end{subfigure}%
\begin{subfigure}[t]{.52\textwidth}
\centering
\begin{tikzpicture}
\begin{axis}[
width=1*\textwidth,
height=.5\textwidth,
xticklabels from table={\textdata}{nsent},
xticklabel style={font=\small},
xtick=data,
nodes near coords align={vertical},
ymin=19,ymax=26,
xlabel={Amount of text data (sentences)},
grid=both,
style=thick
]
\addplot[black] plot[error bars/.cd, y dir = both, y explicit] table[x expr=\coordindex,y=mean,y error=stdv]{\textdata};
\end{axis}
\end{tikzpicture}
\end{subfigure}
\caption{Effect of the amount of unlabeled audio data (left) and text data (right) on unsupervised training in terms of PER on \libri{} dev-other.
}
\label{fig:ablation_data}
\end{figure}
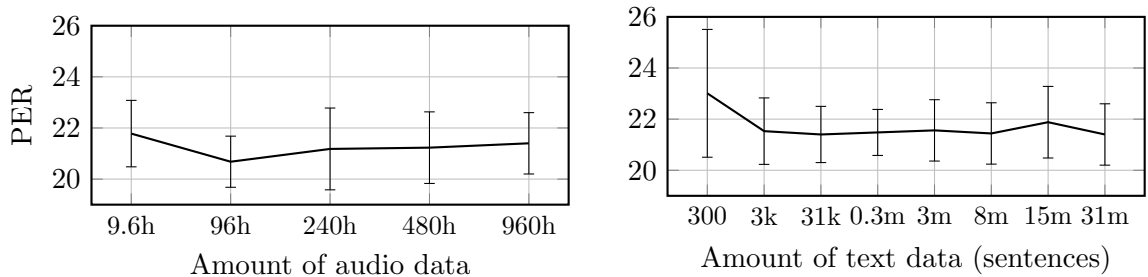

\autoref{fig:ablation_data} (left) shows that 9.6h of speech audio data still achieves excellent performance.
Similarly, \autoref{fig:ablation_data} (right) shows that only about 3,000 sentences of text data are sufficient to achieve a similar level of accuracy as using all of the text data.
We show further ablations in Appendix~\ref{app:ablations}.

\section{Related Work}

This paper builds on a large body of prior work which can be sub-divided into self-supervised learning, semi-supervised learning, unsupervised learning for speech recognition and unsupervised learning for machine translation.

\paragraph{Semi-supervised Speech Recognition.}
This is a long-standing research area in speech recognition with the most recent work focusing on self-training~\citep{kahn2020st} and iterative self-training~\citep{xu2020iterative} by carefully filtering the unlabeled data to match the target domain~\citep{park2020improved}.
There has also been recent work on distilling knowledge from a strong prior such as a language model to provide a learning signal to a discriminative model trained on unlabeled speech~\citep{hsu2020semi}.
Other work explored data augmentation via a technique similar to back-translation~\citep{hayashi2018slt} as used in machine translation~\citep{bojar2011bt_pbmt,sennrich2015improving,edunov2018understanding}.
Another line of work composes text-to-speech and speech-to-text models to enforce cycle-consistency~\citep{hori2019cycle}, similar to dual learning in machine translation~\citep{xia2016dual}.

\paragraph{Self-supervised Learning for Speech.}
Contrastive predictive coding (CPC; \citealt{oord2018cpc}) explored unsupervised or self-supervised representation learning for phoneme recognition. 
Their model architecture and loss function was simplified in~\citet{schneider2019wav2vec} who also applied it to full speech recognition.
\citet{chung2019apc} explores language model-style pre-training on speech audio data with recurrent neural networks and \citet{chung2018speech2vec} learns fixed size representations of audio segments.
Another line of work explored quantization of the continuous speech data to identify speech units in order to learn representations~\citep{baevski2019vqwav2vec,baevski2019effectiveness,alex2019unsupervised,vanniekerk2020vectorquantized,baevski2020wav}. 
\citet{hsu2020hubert} leverages automatically discovered speech units to learn representations with a BERT-like masked prediction objective and achieves comparable results with models based on contrastive learning.
More recently, Transformer architectures have been adopted for self-supervised pre-training~\citep{jiang2019improving,baevski2020wav}. 
There has also been work on robustness of self-supervised learning to domain shift~\citep{hsu2021robust}, multilingual self-supervised speech representation learning~\citep{kawakami2020learning,conneau2020unsupervised} as well as work on combining modalities such as speech and vision~\citep{harwath2019learning}.

\paragraph{Unsupervised Speech Recognition.}
Learning to map speech to phonemes without supervision using adversarial learning has been explored by~\citet{liu2018completely} who learn a mapping matrix between segment identifiers and phonemes. 
However, their work still relied on data segmented into phonemes by human annotators.
This has been later extended to use an automatic segmentation~\citep{chen2019completely} which is iteratively refined with HMMs.
However, cross validation is still performed using labeled data (personal communication with authors).
We also explored HMMs to refine segmentation boundaries but did not find it helpful in our setting. 
Instead, we used HMMs for self-training.

Speech data has been segmented and clustered in a large body of prior work~\citep{varadarajan2008unsupervised,zhang2009unsupervised,gish2009unsupervised,lee2012anb,lee2015unsupervised,ondel2016variational}, including Bayesian methods~\citep{kamper2017seg} and more efficient approximations thereof~\citep{kamper2017embedded}. 
More recent work includes self-supervised approaches to detect phoneme boundaries based on detecting spectral changes in the signal~\citep{kreuk2020self}.

\paragraph{Unsupervised Machine Translation.}
Our work is in part inspired by unsupervised machine translation that showed the possibility of aligning language embedding spaces in an unsupervised way. 
\citet{mikolov2013exploiting} aligned word2vec embedding spaces~\citep{mikolov2013word2vec} by building a linear mapping from a source language to a target language using supervision from a seed dictionary of word translation pairs. 
\cite{artetxe2017learning} greatly reduced the number of word translation pairs needed to learn the linear mapping, by iteratively refining an initial model through semi-supervised learning. 
\cite{conneau2018unsupmt} showed that the supervision used to build the initial alignment could be completely removed by introducing an adversarial game~\citep{goodfellow2014generative,ganin2016domain} in which a generator (the linear mapping from source to target) and a discriminator compete (a classifier trained to identify the languages of randomly sampled word vectors from the projected source space and the target space). 
The generator is trained to fool the discriminator. 
By exploiting the similarties of word2vec spaces across languages, the algorithm converges to a suboptimal mapping that leads to strong unsupervised word translation accuracy. 

Going from unsupervised word translation to unsupervised sentence translation, \cite{lample2018unsupmt,artetxe2018unsupervised} exploited the similarity of neural representations across languages to build fully unsupervised machine translation systems.
They used sequence to sequence architectures~\citep{sutskever2014sequence,bahdanau2014neural} with shared encoder/decoders across languages, and the word embedding layer is initialized with the unsupervised bilingual word embeddings learned in~\citet{conneau2018unsupmt}. 
With this parameter sharing and initialization, a mere bilingual denoising auto-encoding loss provided a first unsupervised machine translation model. 
Similar to word embeddings, semi-supervised learning in the form of back-translation~\citep{sennrich2015improving} further refines this initial model to provide better BLEU scores. 
It was later shown in~\citet{conneau2019cross} that a better self-supervised initialization of the encoder and decoder using cross-lingual masked language modeling~\citep{devlin2018bert} led to significantly better performance. 
Our work follows similar steps than in unsupervised machine translation: pretraining, then unsupervised alignment and finally semi-supervised learning.

\section{Discussion}

\paragraph{Phonemization.}
Our approach requires tools to phonemize text for the language of interest. 
Even for existing tools, we found that the quality of phonemization differs, e.g., G2P vs Phonemizer (\autoref{sec:setup_phonemization}) on the same English data results in a performance difference.
Moreover, phonemizers are not available for all languages and this presents a bottleneck.
To address this, future work may develop phonemizers for more languages, explore phonemization approaches that generalize across languages, or unsupervised training with graphemic text units such as letters. 

\paragraph{Segmentation.}
In our work, we explored a simple segmentation technique based on self-supervised representations, however, there is a large body of work on segmentation and some of these techniques may lead to improvements over our simple approach.
Also, \wvpp{} learns representations for fixed size units with a fixed stride, however, phonemic units are of variable size. Another direction is to learn variable sized representations during pre-training.

\section{Conclusion}

\wvz{} is a framework which enables building speech recognition models without labeled data. 
It embeds and segments the speech audio with self-supervised representations from \wvpp{}, learns a mapping to phonemes with adversarial learning, and cross-validates hyper-parameter choices as well as early stopping with an unsupervised metric. 
Experiments on the standard \libri{} benchmark show performance close to the state of the art models from only a few years ago, even though these models relied on nearly 1,000 hours of labeled data.
Compared to the previous best unsupervised speech recognition approach, \wvz{} reduces TIMIT phoneme error rate from 26.1 to 11.3.
We also demonstrate the viability of our approach on several languages other than English, some of which are low-resource.
The ability to build speech recognition models solely from unlabeled speech audio and unlabeled text drastically lowers the effort to build speech technology for many more languages of the world.


\acks{
We thank Zhouhan Lin for helping with initial explorations in this project, Tatiana Likhomanenko for helpful discussions about self-training, Da-Rong Liu for sharing details to reproduce the setup of~\citet{chen2019completely}, Marc'Aurelio Ranzato for general helpful discussions, and Ruth Kipng'eno, Ruth Ndila Ndeto as well as Mark Mutitu for error analysis of our Swahili model.
}

\newpage
\appendix

\section{Hyperparameter Ablations}
\label{app:ablations}

\insertAblation


\bibliography{refs}

\begin{thebibliography}{127}
\providecommand{\natexlab}[1]{#1}
\providecommand{\url}[1]{\texttt{#1}}
\expandafter\ifx\csname urlstyle\endcsname\relax
  \providecommand{\doi}[1]{doi: #1}\else
  \providecommand{\doi}{doi: \begingroup \urlstyle{rm}\Url}\fi

\bibitem[Abate et~al.(2005)Abate, Menzel, and Tafila]{abate2005amharic}
S.~T. Abate, W.~Menzel, and B.~Tafila.
\newblock An amharic speech corpus for large vocabulary continuous speech
  recognition.
\newblock In \emph{Proc. of Interspeech}, 2005.

\bibitem[Abdel-Hamid et~al.(2012)Abdel-Hamid, Mohamed, Jiang, and
  Penn]{abdel2012applying}
O.~Abdel-Hamid, A.~Mohamed, H.~Jiang, and G.~Penn.
\newblock Applying convolutional neural networks concepts to hybrid nn-hmm
  model for speech recognition.
\newblock In \emph{Proc. of ICASSP}, 2012.

\bibitem[Amodei et~al.(2016)Amodei, Ananthanarayanan, Anubhai, Bai, Battenberg,
  Case, Casper, Catanzaro, Cheng, Chen, et~al.]{amodei2016deepspeech}
D.~Amodei, S.~Ananthanarayanan, R.~Anubhai, J.~Bai, E.~Battenberg, C.~Case,
  J.~Casper, B.~Catanzaro, Q.~Cheng, G.~Chen, et~al.
\newblock Deep speech 2: End-to-end speech recognition in english and mandarin.
\newblock In \emph{Proc. of ICML}, 2016.

\bibitem[Ardila et~al.(2020)Ardila, Branson, Davis, Henretty, Kohler, Meyer,
  Morais, Saunders, Tyers, and Weber]{ardila2019common}
R.~Ardila, M.~Branson, K.~Davis, M.~Henretty, M.~Kohler, J.~Meyer, R.~Morais,
  L.~Saunders, F.~M. Tyers, and G.~Weber.
\newblock Common voice: A massively-multilingual speech corpus.
\newblock \emph{Proc. of LREC}, 2020.

\bibitem[Arjovsky et~al.(2017)Arjovsky, Chintala, and
  Bottou]{arjovsky2017wasserstein}
M.~Arjovsky, S.~Chintala, and L.~Bottou.
\newblock Wasserstein gan.
\newblock \emph{Proc. of ICML}, 2017.

\bibitem[Artetxe et~al.(2017)Artetxe, Labaka, and Agirre]{artetxe2017learning}
M.~Artetxe, G.~Labaka, and E.~Agirre.
\newblock Learning bilingual word embeddings with (almost) no bilingual data.
\newblock In \emph{Proc. of ACL}, 2017.

\bibitem[Artetxe et~al.(2018)Artetxe, Labaka, Agirre, and
  Cho]{artetxe2018unsupervised}
M.~Artetxe, G.~Labaka, E.~Agirre, and K.~Cho.
\newblock Unsupervised neural machine translation.
\newblock \emph{Proc. of ICLR}, 2018.

\bibitem[Baevski and Auli(2018)]{baevski2018adaptive}
A.~Baevski and M.~Auli.
\newblock Adaptive input representations for neural language modeling.
\newblock In \emph{Proc. of ICLR}, 2018.

\bibitem[Baevski et~al.(2020{\natexlab{a}})Baevski, Auli, and
  Mohamed]{baevski2019effectiveness}
A.~Baevski, M.~Auli, and A.~Mohamed.
\newblock Effectiveness of self-supervised pre-training for speech recognition.
\newblock \emph{Proc. of ICASSP}, 2020{\natexlab{a}}.

\bibitem[Baevski et~al.(2020{\natexlab{b}})Baevski, Schneider, and
  Auli]{baevski2019vqwav2vec}
A.~Baevski, S.~Schneider, and M.~Auli.
\newblock vq-wav2vec: Self-supervised learning of discrete speech
  representations.
\newblock In \emph{Proc. of ICLR}, 2020{\natexlab{b}}.

\bibitem[Baevski et~al.(2020{\natexlab{c}})Baevski, Zhou, Mohamed, and
  Auli]{baevski2020wav}
A.~Baevski, Y.~Zhou, A.~Mohamed, and M.~Auli.
\newblock wav2vec 2.0: {A} framework for self-supervised learning of speech
  representations.
\newblock In \emph{Proc. of NeurIPS}, 2020{\natexlab{c}}.

\bibitem[Bahdanau et~al.(2014)Bahdanau, Cho, and Bengio]{bahdanau2014neural}
D.~Bahdanau, K.~Cho, and Y.~Bengio.
\newblock Neural machine translation by jointly learning to align and
  translate.
\newblock \emph{Proc. of ICLR}, 2014.

\bibitem[Bahl et~al.(1986)Bahl, Brown, De~Souza, and Mercer]{bahl1986maximum}
L.~Bahl, P.~Brown, P.~De~Souza, and R.~Mercer.
\newblock Maximum mutual information estimation of hidden markov model
  parameters for speech recognition.
\newblock In \emph{Proc. of ICASSP}, volume~11, pages 49--52. IEEE, 1986.

\bibitem[Bahl et~al.(1983)Bahl, Jelinek, and Mercer]{bahl1983maximum}
L.~R. Bahl, F.~Jelinek, and R.~L. Mercer.
\newblock A maximum likelihood approach to continuous speech recognition.
\newblock \emph{IEEE transactions on pattern analysis and machine
  intelligence}, 1983.

\bibitem[Besacier et~al.(2015)Besacier, Gauthier, Mangeot, Bretier, Bagshaw,
  Rosec, Moudenc, Pellegrino, Voisin, Marsico, and
  Nocera]{besacier2015speechtf}
L.~Besacier, E.~Gauthier, M.~Mangeot, P.~Bretier, P.~Bagshaw, O.~Rosec,
  T.~Moudenc, F.~Pellegrino, S.~Voisin, E.~Marsico, and P.~Nocera.
\newblock Speech technologies for african languages: example of a multilingual
  calculator for education.
\newblock In \emph{Proc. of Interspeech}, 2015.

\bibitem[Bojar and Tamchyna(2011)]{bojar2011bt_pbmt}
O.~Bojar and A.~Tamchyna.
\newblock Improving translation model by monolingual data.
\newblock In \emph{Proc. of WMT}, 2011.

\bibitem[Bourlard and Morgan(2012)]{bourlard2012connectionist}
H.~A. Bourlard and N.~Morgan.
\newblock \emph{Connectionist speech recognition: a hybrid approach}, volume
  247.
\newblock Springer Science \& Business Media, 2012.

\bibitem[Can et~al.(2018)Can, Martinez, Papadopoulos, and
  Narayanan]{can2018pykaldi}
D.~Can, V.~R. Martinez, P.~Papadopoulos, and S.~S. Narayanan.
\newblock Pykaldi: A python wrapper for kaldi.
\newblock In \emph{Proc. of ICASSP}, 2018.

\bibitem[Chen et~al.(2019)Chen, Tsai, Liu, Lee, and shan
  Lee]{chen2019completely}
K.-Y. Chen, C.-P. Tsai, D.-R. Liu, H.-Y. Lee, and L.~shan Lee.
\newblock Completely unsupervised speech recognition by a generative
  adversarial network harmonized with iteratively refined hidden markov models.
\newblock In \emph{Proc. of Interspeech}, 2019.

\bibitem[Chorowski and Jaitly(2017)]{chorowski2016towards}
J.~Chorowski and N.~Jaitly.
\newblock Towards better decoding and language model integration in sequence to
  sequence models.
\newblock \emph{Proc. of Interspeech}, 2017.

\bibitem[Chorowski et~al.(2015)Chorowski, Bahdanau, Serdyuk, Cho, and
  Bengio]{chorowski2015attention}
J.~Chorowski, D.~Bahdanau, D.~Serdyuk, K.~Cho, and Y.~Bengio.
\newblock Attention-based models for speech recognition.
\newblock \emph{Proc. of NIPS}, 2015.

\bibitem[Chung et~al.(2018)Chung, Weng, Tong, and Glass]{chung2018unsup}
Y.~Chung, W.~Weng, S.~Tong, and J.~R. Glass.
\newblock Unsupervised cross-modal alignment of speech and text embedding
  spaces.
\newblock \emph{Proc. of NIPS}, 2018.

\bibitem[Chung et~al.(2019{\natexlab{a}})Chung, Hsu, Tang, and
  Glass]{chung2019apc}
Y.~Chung, W.~Hsu, H.~Tang, and J.~R. Glass.
\newblock An unsupervised autoregressive model for speech representation
  learning.
\newblock \emph{Proc. of Interspeech}, 2019{\natexlab{a}}.

\bibitem[Chung and Glass(2018)]{chung2018speech2vec}
Y.-A. Chung and J.~Glass.
\newblock Speech2vec: A sequence-to-sequence framework for learning word
  embeddings from speech.
\newblock \emph{Proc. of Interspeech}, 2018.

\bibitem[Chung et~al.(2019{\natexlab{b}})Chung, Hsu, Tang, and
  Glass]{chung2019unsupervised}
Y.-A. Chung, W.-N. Hsu, H.~Tang, and J.~Glass.
\newblock An unsupervised autoregressive model for speech representation
  learning.
\newblock \emph{Proc. of Interspeech}, 2019{\natexlab{b}}.

\bibitem[Conneau and Lample(2019)]{conneau2019cross}
A.~Conneau and G.~Lample.
\newblock Cross-lingual language model pretraining.
\newblock \emph{Proc. of NeurIPS}, 2019.

\bibitem[Conneau et~al.(2018)Conneau, Lample, Ranzato, Denoyer, and
  J{\'{e}}gou]{conneau2018unsupmt}
A.~Conneau, G.~Lample, M.~Ranzato, L.~Denoyer, and H.~J{\'{e}}gou.
\newblock Word translation without parallel data.
\newblock \emph{Proc. of ICLR}, 2018.

\bibitem[Conneau et~al.(2020)Conneau, Baevski, Collobert, Mohamed, and
  Auli]{conneau2020unsupervised}
A.~Conneau, A.~Baevski, R.~Collobert, A.~Mohamed, and M.~Auli.
\newblock Unsupervised cross-lingual representation learning for speech
  recognition.
\newblock \emph{arXiv}, abs/2006.13979, 2020.

\bibitem[Devlin et~al.(2019)Devlin, Chang, Lee, and Toutanova]{devlin2018bert}
J.~Devlin, M.-W. Chang, K.~Lee, and K.~Toutanova.
\newblock Bert: Pre-training of deep bidirectional transformers for language
  understanding.
\newblock \emph{Proc. of NAACL}, 2019.

\bibitem[Dieleman et~al.(2018)Dieleman, van~den Oord, and
  Simonyan]{dieleman2018challenge}
S.~Dieleman, A.~van~den Oord, and K.~Simonyan.
\newblock The challenge of realistic music generation: modelling raw audio at
  scale.
\newblock \emph{Proc of NIPS}, 2018.

\bibitem[Dong et~al.(2018)Dong, Xu, and Xu]{linhao2018transformer}
L.~Dong, S.~Xu, and B.~Xu.
\newblock Speech-transformer: A no-recurrence sequence-to-sequence model for
  speech recognition.
\newblock In \emph{Proc. of ICASSP}, 2018.

\bibitem[Edunov et~al.(2018)Edunov, Ott, Auli, and
  Grangier]{edunov2018understanding}
S.~Edunov, M.~Ott, M.~Auli, and D.~Grangier.
\newblock Understanding back-translation at scale.
\newblock In \emph{Proc. of EMNLP}, 2018.

\bibitem[Fan et~al.(2021)Fan, Li, Zhou, and Xu]{fan2021exploring}
Z.~Fan, M.~Li, S.~Zhou, and B.~Xu.
\newblock Exploring wav2vec 2.0 on speaker verification and language
  identification.
\newblock \emph{arXiv}, 2021.

\bibitem[Fer et~al.(2017)Fer, Mat{\v{e}}jka, Gr{\'e}zl, Plchot, Vesel{\`y}, and
  {\v{C}}ernock{\`y}]{fer2017multilingually}
R.~Fer, P.~Mat{\v{e}}jka, F.~Gr{\'e}zl, O.~Plchot, K.~Vesel{\`y}, and J.~H.
  {\v{C}}ernock{\`y}.
\newblock Multilingually trained bottleneck features in spoken language
  recognition.
\newblock \emph{Computer Speech \& Language}, 46, 2017.

\bibitem[Ganin et~al.(2016)Ganin, Ustinova, Ajakan, Germain, Larochelle,
  Laviolette, Marchand, and Lempitsky]{ganin2016domain}
Y.~Ganin, E.~Ustinova, H.~Ajakan, P.~Germain, H.~Larochelle, F.~Laviolette,
  M.~Marchand, and V.~Lempitsky.
\newblock Domain-adversarial training of neural networks.
\newblock \emph{The journal of machine learning research}, 17\penalty0
  (1):\penalty0 2096--2030, 2016.

\bibitem[Garofolo et~al.(1993)Garofolo, Lamel, Fisher, Fiscus, Pallett, and
  Dahlgren]{garofolo1993timit}
J.~S. Garofolo, L.~F. Lamel, W.~M. Fisher, J.~G. Fiscus, D.~S. Pallett, and
  N.~L. Dahlgren.
\newblock {The DARPA TIMIT Acoustic-Phonetic Continuous Speech Corpus CDROM}.
\newblock \emph{Linguistic Data Consortium}, 1993.

\bibitem[Gelas et~al.(2012)Gelas, Besacier, and Pellegrino]{gelas2012alffa}
H.~Gelas, L.~Besacier, and F.~Pellegrino.
\newblock {D}evelopments of {S}wahili resources for an automatic speech
  recognition system.
\newblock In \emph{Proc. of SLTU}, 2012.

\bibitem[Gish et~al.(2009)Gish, Siu, Chan, and Belfield]{gish2009unsupervised}
H.~Gish, M.~Siu, A.~Chan, and W.~Belfield.
\newblock Unsupervised training of an hmm-based speech recognizer for topic
  classification.
\newblock In \emph{Proc. of Interspeech}, 2009.

\bibitem[Goodfellow et~al.(2014)Goodfellow, Pouget-Abadie, Mirza, Xu,
  Warde-Farley, Ozair, Courville, and Bengio]{goodfellow2014generative}
I.~J. Goodfellow, J.~Pouget-Abadie, M.~Mirza, B.~Xu, D.~Warde-Farley, S.~Ozair,
  A.~Courville, and Y.~Bengio.
\newblock Generative adversarial networks.
\newblock \emph{Proc. of NIPS}, 2014.

\bibitem[Google(2021)]{google2021asr}
Google.
\newblock Google cloud: Speech-to-text.
\newblock \url{https://cloud.google.com/speech-to-text}, 2021.
\newblock Accessed: 2021-05-13.

\bibitem[Graves(2012)]{graves2012sequence}
A.~Graves.
\newblock Sequence transduction with recurrent neural networks.
\newblock \emph{Proc. of ICML workshop on Representation Learning}, 2012.

\bibitem[Graves et~al.(2006)Graves, Fernández, and Gomez]{graves2006ctc}
A.~Graves, S.~Fernández, and F.~Gomez.
\newblock Connectionist temporal classification: Labelling unsegmented sequence
  data with recurrent neural networks.
\newblock In \emph{Proc. of ICML}, 2006.

\bibitem[Gulati et~al.(2020)Gulati, Qin, Chiu, Parmar, Zhang, Yu, Han, Wang,
  Zhang, Wu, and Pang]{gulati2020conformer}
A.~Gulati, J.~Qin, C.-C. Chiu, N.~Parmar, Y.~Zhang, J.~Yu, W.~Han, S.~Wang,
  Z.~Zhang, Y.~Wu, and R.~Pang.
\newblock Conformer: Convolution-augmented transformer for speech recognition.
\newblock \emph{Proc. of Interspeech}, 2020.

\bibitem[Gulrajani et~al.(2017)Gulrajani, Ahmed, Arjovsky, Dumoulin, and
  Courville]{gulrajani2017improved}
I.~Gulrajani, F.~Ahmed, M.~Arjovsky, V.~Dumoulin, and A.~Courville.
\newblock Improved training of wasserstein gans.
\newblock \emph{Proc. of NIPS}, 2017.

\bibitem[Han et~al.(2020)Han, Zhang, Zhang, Yu, Chiu, Qin, Gulati, Pang, and
  Wu]{han2020contextnet}
W.~Han, Z.~Zhang, Y.~Zhang, J.~Yu, C.-C. Chiu, J.~Qin, A.~Gulati, R.~Pang, and
  Y.~Wu.
\newblock Contextnet: Improving convolutional neural networks for automatic
  speech recognition with global context.
\newblock \emph{Proc. of Interspeech}, 2020.

\bibitem[Hannun(2017)]{hannun2017sequence}
A.~Hannun.
\newblock Sequence modeling with ctc.
\newblock \emph{Distill}, 2017.
\newblock \doi{10.23915/distill.00008}.
\newblock https://distill.pub/2017/ctc.

\bibitem[Hannun et~al.(2020)Hannun, Pratap, Kahn, and
  Hsu]{hannun2020differentiable}
A.~Hannun, V.~Pratap, J.~Kahn, and W.-N. Hsu.
\newblock Differentiable weighted finite-state transducers.
\newblock \emph{arXiv preprint arXiv:2010.01003}, 2020.

\bibitem[Harwath and Glass(2019)]{harwath2019towards}
D.~Harwath and J.~Glass.
\newblock Towards visually grounded sub-word speech unit discovery.
\newblock In \emph{Proc. of ICASSP}, pages 3017--3021. IEEE, 2019.

\bibitem[Harwath et~al.(2020)Harwath, Hsu, and Glass]{harwath2019learning}
D.~Harwath, W.-N. Hsu, and J.~Glass.
\newblock Learning hierarchical discrete linguistic units from
  visually-grounded speech.
\newblock In \emph{Proc. of ICLR}, 2020.

\bibitem[Hayashi et~al.(2018)Hayashi, Watanabe, Zhang, Toda, Hori, Astudillo,
  and Takeda]{hayashi2018slt}
T.~Hayashi, S.~Watanabe, Y.~Zhang, T.~Toda, T.~Hori, R.~F. Astudillo, and
  K.~Takeda.
\newblock Back-translation-style data augmentation for end-to-end {ASR}.
\newblock In \emph{Proc. of SLT}, 2018.

\bibitem[Heafield(2011)]{heafield-2011-kenlm}
K.~Heafield.
\newblock {K}en{LM}: Faster and smaller language model queries.
\newblock In \emph{Proceedings of the Sixth Workshop on Statistical Machine
  Translation}, pages 187--197, Edinburgh, Scotland, July 2011. Association for
  Computational Linguistics.

\bibitem[Hinton et~al.(2012)Hinton, Deng, Yu, Dahl, Mohamed, Jaitly, Senior,
  Vanhoucke, Nguyen, Sainath, et~al.]{hinton2012deep}
G.~Hinton, L.~Deng, D.~Yu, G.~E. Dahl, A.~Mohamed, N.~Jaitly, A.~Senior,
  V.~Vanhoucke, P.~Nguyen, T.~N. Sainath, et~al.
\newblock Deep neural networks for acoustic modeling in speech recognition: The
  shared views of four research groups.
\newblock \emph{IEEE Signal processing magazine}, 29\penalty0 (6):\penalty0
  82--97, 2012.

\bibitem[Hirsh-Pasek et~al.(1987)Hirsh-Pasek, {Kemler Nelson}, Jusczyk,
  Cassidy, Druss, and Kennedy]{HIRSHPASEK1987269}
K.~Hirsh-Pasek, D.~G. {Kemler Nelson}, P.~W. Jusczyk, K.~W. Cassidy, B.~Druss,
  and L.~Kennedy.
\newblock Clauses are perceptual units for young infants.
\newblock \emph{Cognition}, 26\penalty0 (3):\penalty0 269--286, 1987.

\bibitem[Hori et~al.(2019)Hori, Astudillo, Hayashi, Zhang, Watanabe, and
  Le~Roux]{hori2019cycle}
T.~Hori, R.~Astudillo, T.~Hayashi, Y.~Zhang, S.~Watanabe, and J.~Le~Roux.
\newblock Cycle-consistency training for end-to-end speech recognition.
\newblock In \emph{Proc. of ICASSP}, 2019.

\bibitem[Hsu et~al.(2020)Hsu, Lee, Synnaeve, and Hannun]{hsu2020semi}
W.-N. Hsu, A.~Lee, G.~Synnaeve, and A.~Hannun.
\newblock Semi-supervised speech recognition via local prior matching.
\newblock \emph{arXiv preprint arXiv:2002.10336}, 2020.

\bibitem[Hsu et~al.(2021{\natexlab{a}})Hsu, Sriram, Baevski, Likhomanenko, Xu,
  Pratap, Kahn, Lee, Collobert, Synnaeve, et~al.]{hsu2021robust}
W.-N. Hsu, A.~Sriram, A.~Baevski, T.~Likhomanenko, Q.~Xu, V.~Pratap, J.~Kahn,
  A.~Lee, R.~Collobert, G.~Synnaeve, et~al.
\newblock Robust wav2vec 2.0: Analyzing domain shift in self-supervised
  pre-training.
\newblock \emph{arXiv preprint arXiv:2104.01027}, 2021{\natexlab{a}}.

\bibitem[Hsu et~al.(2021{\natexlab{b}})Hsu, Tsai, Bolte, Salakhutdinov, and
  Mohamed]{hsu2020hubert}
W.-N. Hsu, Y.-H.~H. Tsai, B.~Bolte, R.~Salakhutdinov, and A.~Mohamed.
\newblock Hubert: How much can a bad teacher benefit {ASR} pre-training?
\newblock In \emph{Proc. of ICASSP}, 2021{\natexlab{b}}.

\bibitem[Jang et~al.(2016)Jang, Gu, and Poole]{jang2016gumbel}
E.~Jang, S.~Gu, and B.~Poole.
\newblock Categorical reparameterization with gumbel-softmax.
\newblock \emph{Proc. of ICLR}, 2016.

\bibitem[Jawahar et~al.(2019)Jawahar, Sagot, and Seddah]{jawahar2019does}
G.~Jawahar, B.~Sagot, and D.~Seddah.
\newblock What does bert learn about the structure of language?
\newblock In \emph{Proc. of ACL}, 2019.

\bibitem[Jegou et~al.(2011)Jegou, Douze, and Schmid]{jegou2011ieee}
H.~Jegou, M.~Douze, and C.~Schmid.
\newblock Product quantization for nearest neighbor search.
\newblock \emph{IEEE Trans. Pattern Anal. Mach. Intell.}, 33\penalty0
  (1):\penalty0 117--128, Jan. 2011.

\bibitem[Jiang et~al.(2019)Jiang, Lei, Li, Luo, Hu, Zou, and
  Li]{jiang2019improving}
D.~Jiang, X.~Lei, W.~Li, N.~Luo, Y.~Hu, W.~Zou, and X.~Li.
\newblock Improving transformer-based speech recognition using unsupervised
  pre-training.
\newblock \emph{Proc. of Interspeech}, 2019.

\bibitem[Johnson and Jusczyk(2001)]{JOHNSON2001548}
E.~K. Johnson and P.~W. Jusczyk.
\newblock Word segmentation by 8-month-olds: When speech cues count more than
  statistics.
\newblock \emph{Journal of Memory and Language}, 44\penalty0 (4):\penalty0
  548--567, 2001.

\bibitem[Johnson et~al.(2019)Johnson, Douze, and J{\'e}gou]{johnson2017faiss}
J.~Johnson, M.~Douze, and H.~J{\'e}gou.
\newblock Billion-scale similarity search with gpus.
\newblock \emph{IEEE Transactions on Big Data}, 2019.

\bibitem[Juang et~al.(1986)Juang, Levinson, and Sondhi]{juang1986maximum}
B.-H. Juang, S.~Levinson, and M.~Sondhi.
\newblock Maximum likelihood estimation for multivariate mixture observations
  of markov chains (corresp.).
\newblock \emph{IEEE Transactions on Information Theory}, 32\penalty0
  (2):\penalty0 307--309, 1986.

\bibitem[Jusczyk et~al.(1999)Jusczyk, Houston, and Newsome]{JUSCZYK1999159}
P.~W. Jusczyk, D.~M. Houston, and M.~Newsome.
\newblock The beginnings of word segmentation in english-learning infants.
\newblock \emph{Cognitive Psychology}, 39\penalty0 (3):\penalty0 159--207,
  1999.

\bibitem[Kahn et~al.(2020{\natexlab{a}})Kahn, Lee, and Hannun]{kahn2020st}
J.~Kahn, A.~Lee, and A.~Hannun.
\newblock Self-training for end-to-end speech recognition.
\newblock In \emph{Proc. of ICASSP}, 2020{\natexlab{a}}.

\bibitem[Kahn et~al.(2020{\natexlab{b}})]{kahn2020librilight}
J.~Kahn et~al.
\newblock Libri-light: A benchmark for asr with limited or no supervision.
\newblock In \emph{Proc. of ICASSP}, 2020{\natexlab{b}}.

\bibitem[Kamper et~al.(2017{\natexlab{a}})Kamper, Jansen, and
  Goldwater]{kamper2017seg}
H.~Kamper, A.~Jansen, and S.~Goldwater.
\newblock A segmental framework for fully-unsupervised large-vocabulary speech
  recognition.
\newblock \emph{Comput. Speech Lang.}, 46\penalty0 (C), Nov.
  2017{\natexlab{a}}.

\bibitem[Kamper et~al.(2017{\natexlab{b}})Kamper, Livescu, and
  Goldwater]{kamper2017embedded}
H.~Kamper, K.~Livescu, and S.~Goldwater.
\newblock An embedded segmental k-means model for unsupervised segmentation and
  clustering of speech.
\newblock \emph{Proc. of ASRU}, 2017{\natexlab{b}}.

\bibitem[Kawakami et~al.(2020)Kawakami, Wang, Dyer, Blunsom, and van~den
  Oord]{kawakami2020learning}
K.~Kawakami, L.~Wang, C.~Dyer, P.~Blunsom, and A.~van~den Oord.
\newblock Learning robust and multilingual speech representations.
\newblock \emph{Proc. of EMNLP}, 2020.

\bibitem[Kingma and Ba(2015)]{kingma2015adam}
D.~P. Kingma and J.~Ba.
\newblock {Adam: A Method for Stochastic Optimization}.
\newblock In \emph{Proc. of ICLR}, 2015.

\bibitem[Kreuk et~al.(2020)Kreuk, Keshet, and Adi]{kreuk2020self}
F.~Kreuk, J.~Keshet, and Y.~Adi.
\newblock Self-supervised contrastive learning for unsupervised phoneme
  segmentation.
\newblock \emph{Proc. of Interspeech}, 2020.

\bibitem[Lample et~al.(2018)Lample, Denoyer, and Ranzato]{lample2018unsupmt}
G.~Lample, L.~Denoyer, and M.~Ranzato.
\newblock Unsupervised machine translation using monolingual corpora only.
\newblock In \emph{Proc. of ICLR}, 2018.

\bibitem[Lee and Glass(2012)]{lee2012anb}
C.~Lee and J.~R. Glass.
\newblock A nonparametric bayesian approach to acoustic model discovery.
\newblock In \emph{Proc. of ACL}, 2012.

\bibitem[Lee et~al.(2015)Lee, O'Donnell, and Glass]{lee2015unsupervised}
C.~Lee, T.~J. O'Donnell, and J.~R. Glass.
\newblock Unsupervised lexicon discovery from acoustic input.
\newblock \emph{TACL}, 2015.

\bibitem[Lewis et~al.(2016)Lewis, Simon, and Fennig]{lewis2016ethnologue}
M.~P. Lewis, G.~F. Simon, and C.~D. Fennig.
\newblock Ethnologue: Languages of the world, nineteenth edition.
\newblock Online version: \url{http://www.ethnologue.com}, 2016.

\bibitem[Likhomanenko et~al.(2021)Likhomanenko, Xu, Kahn, Synnaeve, and
  Collobert]{likhomanenko2021slimipl}
T.~Likhomanenko, Q.~Xu, J.~Kahn, G.~Synnaeve, and R.~Collobert.
\newblock slimipl: Language-model-free iterative pseudo-labeling.
\newblock \emph{arXiv}, 2021.

\bibitem[Liu et~al.(2019)Liu, Tu, yi~Lee, and shan Lee]{alex2019unsupervised}
A.~H. Liu, T.~Tu, H.~yi~Lee, and L.~shan Lee.
\newblock Towards unsupervised speech recognition and synthesis with quantized
  speech representation learning.
\newblock \emph{Proc. of ICASSP}, 2019.

\bibitem[Liu et~al.(2018)Liu, Chen, Lee, and shan Lee]{liu2018completely}
D.-R. Liu, K.-Y. Chen, H.-Y. Lee, and L.~shan Lee.
\newblock Completely unsupervised phoneme recognition by adversarially learning
  mapping relationships from audio embeddings.
\newblock \emph{Proc. of Interspeech}, 2018.

\bibitem[Manohar et~al.(2018)Manohar, Hadian, Povey, and
  Khudanpur]{manohar2018semi}
V.~Manohar, H.~Hadian, D.~Povey, and S.~Khudanpur.
\newblock Semi-supervised training of acoustic models using lattice-free mmi.
\newblock In \emph{Proc. of ICASSP}, 2018.

\bibitem[Mikolov et~al.(2013{\natexlab{a}})Mikolov, Le, and
  Sutskever]{mikolov2013exploiting}
T.~Mikolov, Q.~V. Le, and I.~Sutskever.
\newblock Exploiting similarities among languages for machine translation.
\newblock \emph{arXiv preprint arXiv:1309.4168}, 2013{\natexlab{a}}.

\bibitem[Mikolov et~al.(2013{\natexlab{b}})Mikolov, Sutskever, Chen, Corrado,
  and Dean]{mikolov2013word2vec}
T.~Mikolov, I.~Sutskever, K.~Chen, G.~S. Corrado, and J.~Dean.
\newblock Distributed representations of words and phrases and their
  compositionality.
\newblock In \emph{Proc. of NIPS}, 2013{\natexlab{b}}.

\bibitem[Mohri(1997)]{mohri1997finite}
M.~Mohri.
\newblock Finite-state transducers in language and speech processing.
\newblock \emph{Computational linguistics}, 23\penalty0 (2):\penalty0 269--311,
  1997.

\bibitem[Mohri et~al.(2002)Mohri, Pereira, and Riley]{mohri2002weighted}
M.~Mohri, F.~Pereira, and M.~Riley.
\newblock Weighted finite-state transducers in speech recognition.
\newblock \emph{Computer Speech \& Language}, 16\penalty0 (1):\penalty0 69--88,
  2002.

\bibitem[Ondel et~al.(2016)Ondel, Burget, and
  Cernock{\'y}]{ondel2016variational}
L.~Ondel, L.~Burget, and J.~Cernock{\'y}.
\newblock Variational inference for acoustic unit discovery.
\newblock In \emph{Proc. of SLTU}, 2016.

\bibitem[Ott et~al.(2019)Ott, Edunov, Baevski, Fan, Gross, Ng, Grangier, and
  Auli]{ott2019fairseq}
M.~Ott, S.~Edunov, A.~Baevski, A.~Fan, S.~Gross, N.~Ng, D.~Grangier, and
  M.~Auli.
\newblock fairseq: A fast, extensible toolkit for sequence modeling.
\newblock In \emph{Proc. of NAACL System Demonstrations}, 2019.

\bibitem[Panayotov et~al.(2015)Panayotov, Chen, Povey, and
  Khudanpur]{panayotov2015librispeech}
V.~Panayotov, G.~Chen, D.~Povey, and S.~Khudanpur.
\newblock Librispeech: an asr corpus based on public domain audio books.
\newblock In \emph{Proc. of ICASSP}, pages 5206--5210. IEEE, 2015.

\bibitem[Park et~al.(2019)Park, Chan, Zhang, Chiu, Zoph, Cubuk, and
  Le]{park2019specaugment}
D.~S. Park, W.~Chan, Y.~Zhang, C.-C. Chiu, B.~Zoph, E.~D. Cubuk, and Q.~V. Le.
\newblock Specaugment: A simple data augmentation method for automatic speech
  recognition.
\newblock In \emph{Proc. of Interspeech}, 2019.

\bibitem[Park et~al.(2020)Park, Zhang, Jia, Han, Chiu, Li, Wu, and
  Le]{park2020improved}
D.~S. Park, Y.~Zhang, Y.~Jia, W.~Han, C.-C. Chiu, B.~Li, Y.~Wu, and Q.~V. Le.
\newblock Improved noisy student training for automatic speech recognition.
\newblock \emph{Proc. of Interspeech}, 2020.

\bibitem[Park and Kim(2019)]{g2pE2019}
K.~Park and J.~Kim.
\newblock g2pe.
\newblock \url{https://github.com/Kyubyong/g2p}, 2019.

\bibitem[Pepino et~al.(2021)Pepino, Riera, and Ferrer]{pepino2021emotion}
L.~Pepino, P.~Riera, and L.~Ferrer.
\newblock Emotion recognition from speech using wav2vec 2.0 embeddings.
\newblock \emph{arXiv}, 2021.

\bibitem[Polka and Werker(1994)]{polka1994developmental}
L.~Polka and J.~F. Werker.
\newblock Developmental changes in perception of nonnative vowel contrasts.
\newblock \emph{Journal of Experimental Psychology: Human perception and
  performance}, 20\penalty0 (2):\penalty0 421, 1994.

\bibitem[Povey(2005)]{povey2005discriminative}
D.~Povey.
\newblock \emph{Discriminative training for large vocabulary speech
  recognition}.
\newblock PhD thesis, University of Cambridge, 2005.

\bibitem[Povey et~al.(2011)Povey, Ghoshal, Boulianne, Burget, Glembek, Goel,
  Hannemann, Motlicek, Qian, Schwarz, Silovsky, Stemmer, and
  Vesely]{povey2011kaldi}
D.~Povey, A.~Ghoshal, G.~Boulianne, L.~Burget, O.~Glembek, N.~Goel,
  M.~Hannemann, P.~Motlicek, Y.~Qian, P.~Schwarz, J.~Silovsky, G.~Stemmer, and
  K.~Vesely.
\newblock The kaldi speech recognition toolkit.
\newblock In \emph{Proc. of ASRU}, 2011.

\bibitem[{Pratap} et~al.(2019){Pratap}, {Hannun}, {Xu}, {Cai}, {Kahn},
  {Synnaeve}, {Liptchinsky}, and {Collobert}]{pratap2019w2l}
V.~{Pratap}, A.~{Hannun}, Q.~{Xu}, J.~{Cai}, J.~{Kahn}, G.~{Synnaeve},
  V.~{Liptchinsky}, and R.~{Collobert}.
\newblock Wav2letter++: A fast open-source speech recognition system.
\newblock In \emph{Proc. of ICASSP}, 2019.

\bibitem[Pratap et~al.(2020)Pratap, Xu, Sriram, Synnaeve, and
  Collobert]{pratap2020mls}
V.~Pratap, Q.~Xu, A.~Sriram, G.~Synnaeve, and R.~Collobert.
\newblock Mls: A large-scale multilingual dataset for speech research.
\newblock In \emph{Proc. of Interspeech}, 2020.

\bibitem[Rao et~al.(2017)Rao, Sak, and Prabhavalkar]{rao2017exploring}
K.~Rao, H.~Sak, and R.~Prabhavalkar.
\newblock Exploring architectures, data and units for streaming end-to-end
  speech recognition with rnn-transducer.
\newblock In \emph{2017 IEEE Automatic Speech Recognition and Understanding
  Workshop (ASRU)}, pages 193--199. IEEE, 2017.

\bibitem[Rasanen et~al.(2015)Rasanen, Doyle, and Frank]{rasanen2015interspeech}
O.~Rasanen, G.~Doyle, and M.~C. Frank.
\newblock Unsupervised word discovery from speech using automatic segmentation
  into syllable-like units.
\newblock In \emph{Proc. of Interspeech}, 2015.

\bibitem[Ravanelli et~al.(2018)Ravanelli, Brakel, Omologo, and
  Bengio]{ravanelli2018lgru}
M.~Ravanelli, P.~Brakel, M.~Omologo, and Y.~Bengio.
\newblock Light gated recurrent units for speech recognition.
\newblock \emph{IEEE Trans. on Emerging Topics in Comp. Intel.}, 2, 2018.

\bibitem[Ravanelli et~al.(2019)Ravanelli, Parcollet, and
  Bengio]{ravanelli2019pytorchkaldi}
M.~Ravanelli, T.~Parcollet, and Y.~Bengio.
\newblock The pytorch-kaldi speech recognition toolkit.
\newblock \emph{Proc. of ICASSP}, 2019.

\bibitem[Rivière et~al.(2020)Rivière, Joulin, Mazaré, and
  Dupoux]{rivire2020unsupervised}
M.~Rivière, A.~Joulin, P.-E. Mazaré, and E.~Dupoux.
\newblock Unsupervised pretraining transfers well across languages.
\newblock In \emph{Proc. of ICASSP}, 2020.

\bibitem[Schneider et~al.(2019)Schneider, Baevski, Collobert, and
  Auli]{schneider2019wav2vec}
S.~Schneider, A.~Baevski, R.~Collobert, and M.~Auli.
\newblock wav2vec: Unsupervised pre-training for speech recognition.
\newblock In \emph{Proc. of Interspeech}, 2019.

\bibitem[Sennrich et~al.(2015)Sennrich, Haddow, and
  Birch]{sennrich2015improving}
R.~Sennrich, B.~Haddow, and A.~Birch.
\newblock Improving neural machine translation models with monolingual data.
\newblock \emph{Proc. of ACL}, 2015.

\bibitem[Srivastava et~al.(2014)Srivastava, Hinton, Krizhevsky, Sutskever, and
  Salakhutdinov]{srivastava2014dropout}
N.~Srivastava, G.~Hinton, A.~Krizhevsky, I.~Sutskever, and R.~Salakhutdinov.
\newblock Dropout: a simple way to prevent neural networks from overfitting.
\newblock \emph{JMLR}, 2014.

\bibitem[Sutskever et~al.(2014)Sutskever, Vinyals, and
  Le]{sutskever2014sequence}
I.~Sutskever, O.~Vinyals, and Q.~V. Le.
\newblock Sequence to sequence learning with neural networks.
\newblock \emph{Proc. of NIPS}, 2014.

\bibitem[Synnaeve et~al.(2020)Synnaeve, Xu, Kahn, Likhomanenko, Grave, Pratap,
  Sriram, Liptchinsky, and Collobert]{synnaeve2020end}
G.~Synnaeve, Q.~Xu, J.~Kahn, T.~Likhomanenko, E.~Grave, V.~Pratap, A.~Sriram,
  V.~Liptchinsky, and R.~Collobert.
\newblock End-to-end {ASR}: from {Supervised} to {Semi}-{Supervised} {Learning}
  with {Modern} {Architectures}.
\newblock \emph{Proc. of ICML workshop on Self-supervision in Audio and Speech
  (SAS)}, 2020.

\bibitem[Tachbelie et~al.(2014)Tachbelie, Abate, and
  Besacier]{tachbelie2014alffa}
M.~Tachbelie, S.~T. Abate, and L.~Besacier.
\newblock Using different acoustic, lexical and language modeling units for asr
  of an under-resourced language - amharic.
\newblock \emph{Speech Communication}, 56, 2014.

\bibitem[Tan et~al.(2020)Tan, Sarkar, and Dehak]{tan_rvad}
Z.~Tan, A.~K. Sarkar, and N.~Dehak.
\newblock rvad: An unsupervised segment-based robust voice activity detection
  method.
\newblock \emph{Computer speech \& language}, 59:\penalty0 1--21, 2020.

\bibitem[Tenney et~al.(2019)Tenney, Das, and Pavlick]{tenney2019bert}
I.~Tenney, D.~Das, and E.~Pavlick.
\newblock Bert rediscovers the classical nlp pipeline.
\newblock \emph{Proc. of ACL}, 2019.

\bibitem[van~den Oord et~al.(2018)van~den Oord, Li, and Vinyals]{oord2018cpc}
A.~van~den Oord, Y.~Li, and O.~Vinyals.
\newblock Representation learning with contrastive predictive coding.
\newblock \emph{Proc. of NIPS}, 2018.

\bibitem[van Niekerk et~al.(2020)van Niekerk, Nortje, and
  Kamper]{vanniekerk2020vectorquantized}
B.~van Niekerk, L.~Nortje, and H.~Kamper.
\newblock Vector-quantized neural networks for acoustic unit discovery in the
  zerospeech 2020 challenge.
\newblock \emph{Proc. of Interspeech}, 2020.

\bibitem[Varadarajan et~al.(2008)Varadarajan, Khudanpur, and
  Dupoux]{varadarajan2008unsupervised}
B.~Varadarajan, S.~Khudanpur, and E.~Dupoux.
\newblock Unsupervised learning of acoustic sub-word units.
\newblock In \emph{Proc. of ACL}, 2008.

\bibitem[Vaswani et~al.(2017)Vaswani, Shazeer, Parmar, Uszkoreit, Jones, Gomez,
  Kaiser, and Polosukhin]{vaswani2017transformer}
A.~Vaswani, N.~Shazeer, N.~Parmar, J.~Uszkoreit, L.~Jones, A.~N. Gomez,
  L.~Kaiser, and I.~Polosukhin.
\newblock Attention is all you need.
\newblock In \emph{Proc. of NIPS}, 2017.

\bibitem[Vesel{\`y} et~al.(2017)Vesel{\`y}, Burget, and
  Cernock{\`y}]{vesely2017semi}
K.~Vesel{\`y}, L.~Burget, and J.~Cernock{\`y}.
\newblock Semi-supervised dnn training with word selection for asr.
\newblock In \emph{Proc. of Interspeech}, 2017.

\bibitem[Wang et~al.(2021)Wang, Wu, Pino, Baevski, Auli, and
  Conneau]{wang2021st}
C.~Wang, A.~Wu, J.~Pino, A.~Baevski, M.~Auli, and A.~Conneau.
\newblock Large-scale self- and semi-supervised learning for speech
  translation.
\newblock \emph{arXiv}, 2021.

\bibitem[Werker and Tees(1984)]{werker1984cross}
J.~F. Werker and R.~C. Tees.
\newblock Cross-language speech perception: Evidence for perceptual
  reorganization during the first year of life.
\newblock \emph{Infant behavior and development}, 7\penalty0 (1):\penalty0
  49--63, 1984.

\bibitem[Xia et~al.(2016)Xia, He, Qin, Wang, Yu, Liu, and Ma]{xia2016dual}
Y.~Xia, D.~He, T.~Qin, L.~Wang, N.~Yu, T.~Liu, and W.~Ma.
\newblock Dual learning for machine translation.
\newblock In \emph{Proc. of NeurIPS}, 2016.

\bibitem[{Xu} et~al.(2018){Xu}, {Li}, {Wang}, {Wang}, {Kang}, {Chen}, {Povey},
  and {Khudanpur}]{xu2018icassp}
H.~{Xu}, K.~{Li}, Y.~{Wang}, J.~{Wang}, S.~{Kang}, X.~{Chen}, D.~{Povey}, and
  S.~{Khudanpur}.
\newblock Neural network language modeling with letter-based features and
  importance sampling.
\newblock In \emph{Proc. of ICASSP}, 2018.

\bibitem[Xu et~al.(2020{\natexlab{a}})Xu, Baevski, Likhomanenko, Tomasello,
  Conneau, Collobert, Synnaeve, and Auli]{xu2020selftraining}
Q.~Xu, A.~Baevski, T.~Likhomanenko, P.~Tomasello, A.~Conneau, R.~Collobert,
  G.~Synnaeve, and M.~Auli.
\newblock Self-training and pre-training are complementary for speech
  recognition.
\newblock In \emph{Proc. of ICASSP}, 2020{\natexlab{a}}.

\bibitem[Xu et~al.(2020{\natexlab{b}})Xu, Likhomanenko, Kahn, Hannun, Synnaeve,
  and Collobert]{xu2020iterative}
Q.~Xu, T.~Likhomanenko, J.~Kahn, A.~Hannun, G.~Synnaeve, and R.~Collobert.
\newblock Iterative pseudo-labeling for speech recognition.
\newblock \emph{Proc. of Interspeech}, 2020{\natexlab{b}}.

\bibitem[Yeh et~al.(2019)Yeh, Chen, Yu, and Yu]{yeh2018unsupervised}
C.-K. Yeh, J.~Chen, C.~Yu, and D.~Yu.
\newblock Unsupervised speech recognition via segmental empirical output
  distribution matching.
\newblock In \emph{Proc. of ICLR}, 2019.

\bibitem[Young(1996)]{young1996large}
S.~Young.
\newblock Large vocabulary continuous speech recognition: A review.
\newblock \emph{IEEE Signal Processing Magazine}, 13\penalty0 (5):\penalty0
  45--57, 1996.

\bibitem[Zeghidour et~al.(2018)Zeghidour, Xu, Liptchinsky, Usunier, Synnaeve,
  and Collobert]{zeghidour2018w2l}
N.~Zeghidour, Q.~Xu, V.~Liptchinsky, N.~Usunier, G.~Synnaeve, and R.~Collobert.
\newblock Fully convolutional speech recognition.
\newblock \emph{arXiv}, abs/1812.06864, 2018.

\bibitem[Zeyer et~al.(2017)Zeyer, Beck, Schl{\"u}ter, and Ney]{zeyer2017ctc}
A.~Zeyer, E.~Beck, R.~Schl{\"u}ter, and H.~Ney.
\newblock Ctc in the context of generalized full-sum hmm training.
\newblock In \emph{Proc. of Interspeech}, 2017.

\bibitem[Zhang et~al.(2020{\natexlab{a}})Zhang, Wang, Zhang, Liu, Saraf, and
  Zweig]{zhang2020faster}
F.~Zhang, Y.~Wang, X.~Zhang, C.~Liu, Y.~Saraf, and G.~Zweig.
\newblock Faster, simpler and more accurate hybrid asr systems using
  wordpieces.
\newblock \emph{Proc. of Interspeech}, 2020{\natexlab{a}}.

\bibitem[Zhang and Glass(2009)]{zhang2009unsupervised}
Y.~Zhang and J.~R. Glass.
\newblock Unsupervised spoken keyword spotting via segmental dtw on gaussian
  posteriorgrams.
\newblock \emph{IEEE Workshop on Automatic Speech Recognition \&
  Understanding}, 2009.

\bibitem[Zhang et~al.(2020{\natexlab{b}})Zhang, Qin, Park, Han, Chiu, Pang, Le,
  and Wu]{zhang2020pushing}
Y.~Zhang, J.~Qin, D.~S. Park, W.~Han, C.-C. Chiu, R.~Pang, Q.~V. Le, and Y.~Wu.
\newblock Pushing the limits of semi-supervised learning for automatic speech
  recognition.
\newblock \emph{Proc. of NeurIPS SAS Workshop}, 2020{\natexlab{b}}.

\end{thebibliography}

\end{document}